\documentclass[]{fairmeta}

\usepackage[utf8]{inputenc}
\usepackage{url}
\usepackage{amsfonts}
\usepackage{nicefrac}
\usepackage{algorithm}
\usepackage{algorithmic}
\usepackage{amsmath}
\usepackage{amssymb}
\usepackage{mathtools}
\usepackage{amsthm}
\usepackage{wrapfig}

\theoremstyle{plain}

\theoremstyle{definition}

\theoremstyle{remark}

\setlabdisplayname{OmniAl Group of ZJU ACES Lab}
\setuniversityname{}

\title{Embodied-Navigator: Point, Think, Memorize, and Align for Efficient Navigation}

\author[* 1]{Hongyan Feng}
\author[* 1]{Sunlai Chen}
\author[* 1]{Xuanyu Liu}
\author[1]{Miao Pan}
\author[1]{Yangfan Xie}
\author[2]{Yuxiang Cui}
\author[2]{Zhongxiang Zhou}
\author[2]{Rong Xiong}
\author[1]{Wenqi Zhang}
\author[1]{Jianwei Yin}
\author[1]{Yueting Zhuang}
\author[\dagger 1]{Xuhong Zhang}

\affiliation[1]{School of Software Technolog,Zhejiang University\\}
\affiliation[2]{Zhejiang Humanoid Robot Innovation Center Co., Ltd.\\}
\contribution[*]{Equal contribution}
\contribution[\dagger]{Corresponding author}
\correspondence{\texttt{zhangxuhong@zju.edu.cn}}

\abstract{
Although Large Vision-Language Models (VLMs) have significantly advanced embodied navigation, their direct deployment remains challenging, as existing methods often force VLMs into unnatural action spaces that misalign with their 2D pre-training priors, compounded by rigid reasoning schedules and inefficient memory management. To overcome these limitations, we propose \textbf{TAMP-Nav}, a unified framework for efficient embodied navigation. First, we introduce a \textbf{Pixel-to-3D Action Formulation} (\textit{Point}) that reformulates navigation into 2D visual prompting. Specifically, the VLM merely selects 2D pixels, which are then projected into 3D coordinates for a low-level SLAM controller. This design naturally aligns embodied execution with the VLM's inherent 2D visual capabilities. Second, we propose an integrated \textbf{Selective Reasoning and Anchor-Trajectory Memory} mechanism (\textit{Think} and \textit{Memorize}), which dynamically triggers Chain-of-Thought and retains high-fidelity memory only at critical nodes, compressing redundant trajectories into lightweight \textbf{Space-Time Indicators}, thereby preserving critical historical information and enhancing spatio-temporal perception. Finally, we design an efficient \textbf{Two-Level Alignment Paradigm} (\textit{Align}) via Group Relative Policy Optimization (GRPO). By superimposing global outcome rewards with fine-grained process rewards, this dense supervision tightly aligns the agent's cognitive planning with physical environmental feedback, endowing the model with adaptive reasoning capabilities. Experiments demonstrate that TAMP-Nav achieves state-of-the-art performance (e.g., 66.2\% SR on R2R-CE) with high runtime and sample efficiency (requiring only 90k training trajectories).
\par\medskip\noindent\textbf{GitHub:} \href{https://github.com/ZJU-OmniAI/Embodied-Omni}{\texttt{ZJU-OmniAI/Embodied-Omni}}
}

\date{August 2026}

\begin{document}
\thispagestyle{firstheader}
\maketitle
\pagestyle{empty}
\section{Introduction}
\label{sec:intro}

Embodied Navigation requires agents to interpret natural language instructions and navigate through complex environments (\cite{zhang2024vision, krantz_vlnce_2020, ku2020room}). While Large Vision-Language Models (VLMs) have vastly improved multi-modal understanding, directly deploying them in real-world Embodied Navigation exposes three critical technical bottlenecks:

\textbf{The Geometric Gap in Action Formulation.} Pioneering VLN methods have made great strides by mapping observations directly to low-level atomic actions (e.g., ``turn left $30^{\circ}$'') (\cite{zhang2024navid,zhang2024uni,cheng2024navila,wei2025streamvln}) or regressing 3D spatial coordinates (\cite{wei2025ground, zhang2025embodied}). However, VLMs are predominantly pre-trained on 2D image-text pairs. Forcing them to implicitly learn complex 3D geometric transformations leads to spatial hallucinations and low sample efficiency (\cite{zhang2025flatland,stogiannidis2025mind}).

\textbf{The Dilemma Between Navigation Performance and Inference Efficiency.} Integrating Chain-of-Thought (CoT) (\cite{wei2022cot}) has enhanced the high-level planning capabilities of navigation models. However, existing models typically trigger reasoning at \textbf{every step or predefined intervals} (\cite{lin2025navcot,song2023llmplanner,liu2026span}). While dense reasoning improves decision-making quality, it also introduces severe inference latency. Ultimately, such rigid cognitive designs fail to strike an optimal balance between runtime efficiency and navigation performance.

\textbf{The Loss of Key Information and Spatio-Temporal Perception in Long-Horizon Memory.} Existing navigation memory mechanisms struggle to manage long-horizon histories effectively. Retaining full historical visual features inevitably leads to attention dilution and memory overflow; conversely, indiscriminately discarding past observations (\cite{zhang2024navid},\cite{wei2025ground}) or forgetting curves (\cite{zhang2025embodied}) may result in the loss of critical information. More importantly, because existing architectures often lack explicit spatio-temporal coordinates to organize this historical information, agents struggle to maintain a clear comprehension of their past trajectories.

To break these bottlenecks, we propose \textbf{TAMP-Nav} (\textbf{T}hink, \textbf{A}lign, \textbf{M}emorize, and \textbf{P}oint).

First, to bridge the geometric gap, we introduce a \textbf{Pixel-to-3D Action Space} (\textit{Point}). To bypass complex 3D reasoning and directly leverage 2D image-text priors, the agent acts as a ``pointer,'' directly outputting 2D pixel coordinates on the egocentric visual observations, which are then projected into 3D and executed by a local SLAM controller (\cite{yang2022far, xu2022fastlio2}).

Second, for efficient reasoning and memorization, we propose an integrated \textbf{Selective Reasoning and Anchor-Trajectory Memory} (\textit{Think and Memorize}). TAMP-Nav discards rigid step-by-step CoT. Instead, the model dynamically triggers deep CoT only at critical topological nodes, which are subsequently stored as high-fidelity \textbf{Explicit Anchors}. Conversely, routine path intervals are bypassed without reasoning and compressed into lightweight \textbf{Space-Time Indicators}, strictly controlling the memory budget while preserving key information and spatio-temporal perception.

Crucially, to more efficiently enhance the model's navigation and adaptive reasoning capabilities, we design a \textbf{Two-Level GRPO} framework (\textit{Align}) featuring an innovative multi-branch rollout mechanism. Moving beyond strict expert-forcing, our method performs autonomous group rollouts: the agent samples a diverse set of candidate actions at each decision node and unrolls them into multiple complete, distinct trajectories. Across these diverse rollouts, we seamlessly superimpose \textbf{global outcome rewards} (task success, trajectory efficiency via SPL, and reasoning density constraints) with explicit, \textbf{fine-grained process rewards} (target approach metrics, collision avoidance, stop action correctness, reasoning value assessment, and format adherence). By evaluating group-relative advantages at both the trajectory and step levels, this dual-reward architecture provides dense supervision that tightly aligns the agent's internal cognitive planning with physical environmental feedback. Through this alignment process, the agent autonomously acquires the meta-cognitive ability to trigger CoT reasoning in a state-dependent manner.

Our main contributions are summarized as follows:

\textbf{(1) Vision-Centric Action Formulation(Point)}: We introduce a Pixel-to-3D action space that decouples high-level visual prompting from low-level execution. This effectively bridges the geometric gap, enabling the agent to leverage 2D VLM pre-training priors for 3D embodied navigation.

\textbf{(2) Efficient Cognitive Architecture (Think and Memorize)}: We propose an integrated mechanism that couples dynamic selective reasoning with Anchor-Trajectory Memory. This resolves the trade-off between inference efficiency and navigation performance while enabling efficient history storage through explicit spatio-temporal encoding and key-node memory.

\textbf{(3) Efficient Two-Level Alignment Paradigm (Align)}: We design a two-level GRPO framework that superimposes global outcome rewards with fine-grained process rewards. This dense supervision tightly aligns the agent's cognitive planning with physical environmental feedback, endowing the model with adaptive reasoning capabilities and achieving state-of-the-art performance (e.g., 66.2\% SR on R2R-CE) with high sample efficiency (requiring only 90k training trajectories).

\section{MultiNav-CoT Dataset}
\label{sec:dataset}

To endow TAMP-Nav with high-level logical reasoning and spatial understanding, we construct \textbf{MultiNav-CoT}, a curated dataset containing 90k trajectories. Derived from VLN-CE (\cite{krantz_vlnce_2020}), these samples are filtered to exclude rendering artifacts (e.g., black images) and Gemini-assessed low-quality instructions. MultiNav-CoT features \textbf{selective Chain-of-Thought} annotations to supervise sparse reasoning and solve the RL cold-start problem. To generate this high-value reasoning, we developed a CoT engine comprising \textbf{Key Node Mining} and \textbf{Structured Generation}.

\textbf{(1) Spatiotemporal Key Node Mining}
To determine exactly \textit{when} the agent should trigger deep reasoning, we propose a three-step Spatiotemporal Key Node Mining strategy (detailed pseudocode is provided in Appendix \ref{sec:appendix_key_node}). 
First, we compute an importance score 
\begin{equation}
\label{eq:mining}
S(t) = S_{sem}(t) + S_{vis}(t)
\end{equation}
For each frame, $S_{sem}$ captures the semantic relevance between the current visual observation and the language instruction via CLIP similarity (\cite{radford2021learning}), while $S_{vis}$ measures scene transitions based on visual feature differences (\cite{dosovitskiy2020image}; detailed formulations are provided in Appendix \ref{sec:appendix_key_node}). Second, we apply a \textbf{distance-based greedy filtering} to the score-sorted frames, iteratively selecting high-scoring nodes while discarding candidates within a minimum spatial distance $D_{min}$ to ensure reasoning sparsity. Third, to maintain global topological connectivity, we perform temporal padding by inserting the highest-scoring intermediate frame whenever the spatial gap between consecutive nodes exceeds $D_{max}$. Ultimately, this dynamic selection yields a sparse set of key nodes comprising roughly 30\% of the trajectory, serving as anchors for CoT generation.

\textbf{(2) Structured CoT Generation}: We adopt a multi-stage prompt engineering pipeline based on \textbf{Gemini 2.5 flash} (\cite{comanici2025gemini}) to construct the reasoning annotations. Rather than prompting the model to generate the entire reasoning chain in a single pass, we decompose the generation into three independent sub-tasks: \textbf{Task Phase Localization}, \textbf{Current Observation Analysis}, and \textbf{Future Action Reasoning}. Generating these components separately reduces the cognitive load of a single LLM query, minimizing hallucinations and ensuring high stability and quality of the CoT. Subsequently, a final fusion step synthesizes these outputs into a unified, coherent narrative. This multi-stage approach establishes a strict logical flow that tightly couples visual perception with long-horizon planning. Detailed prompts are provided in Appendix \ref{sec:appendix_cot_prompts}.

\section{Method}
\label{sec:method}

\begin{figure}[t]
    \centering
    \includegraphics[width=\textwidth]{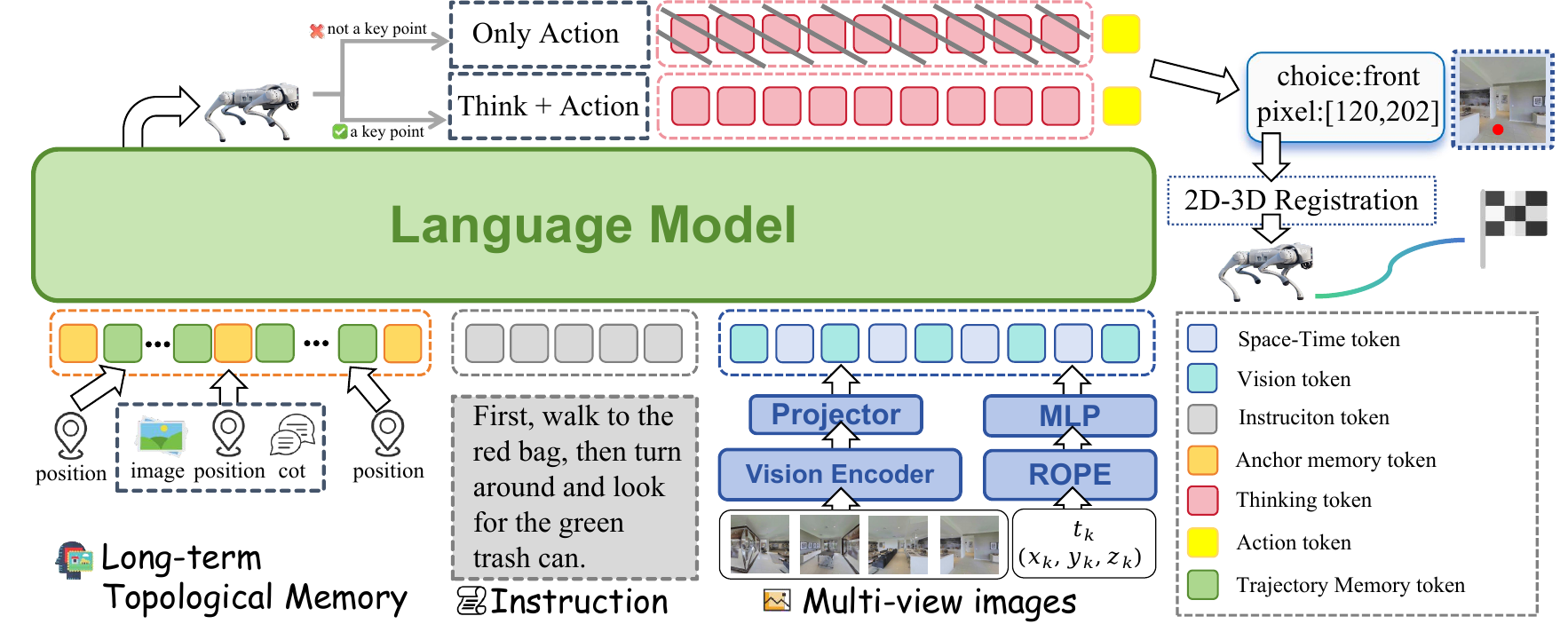}
    \caption{The architecture of TAMP-Nav. Given visual-textual inputs, the agent compresses long-horizon history into an Anchor-Trajectory Memory, autonomously triggering reasoning at critical nodes. As a visual pointer, it selects the optimal multi-camera view and predicts a 2D pixel, which is projected into 3D for SLAM execution. Finally, the policy is optimized via Two-Level GRPO.}
    \label{fig:architecture}
\end{figure}

To seamlessly bridge high-level reasoning with low-level execution while achieving high sample and inference efficiency, we propose \textbf{TAMP-Nav}, as illustrated in Figure \ref{fig:architecture}. Our framework operates through a continuous interaction loop. At each navigation step, the agent first integrates current visual observations with an \textbf{Anchor-Trajectory Memory} to maintain long-horizon spatial awareness. Then, it autonomously decides whether to trigger \textbf{Selective Reasoning} to analyze complex scenarios. Following this, the model employs a \textbf{Pixel-to-3D Action Formulation} to directly predict a pixel waypoint, which is then converted into 3D spatial coordinates and passed to the low-level planner for execution. The entire decision-making process is optimized via \textbf{Two-Level GRPO}, aligning the agent to maximize navigation success while minimizing computational overhead.

\subsection{Pixel-to-3D Action Formulation}
\label{subsec:action_space}

TAMP-Nav acts as a visual pointer. At step $t$, the agent receives four egocentric observations $V_t = \{v_{t,1}, v_{t,2}, v_{t,3}, v_{t,4}\}$ to cover a $360^{\circ}$ field of view. The VLM first selects the optimal view $v_{t,i}$ and subsequently outputs a 2D pixel $a_t = (u, v)$ pointing to the target waypoint. This pixel is projected into a local 3D point $P_t$:
\begin{equation}
    P_t = D_{t,i}(u, v) \cdot K^{-1} [u, v, 1]^T
\end{equation}
where $D_{t,i}$ denotes the depth map and $K$ represents the camera intrinsic matrix. This 3D coordinate is then transformed into the world coordinate system and dispatched to a low-level SLAM controller, which handles local movement to reach $P_t$. This formulation relieves the VLM from learning complex geometric transformations, allowing it to focus entirely on visual-semantic grounding.

In simulation environments, the ground-truth depth map $D_{t,i}$ is directly accessible. For real-world deployment, we directly obtain the depth map using a depth camera. We evaluate robustness to depth inaccuracies in Section \ref{subsec:depth_robustness}. Because navigation operates as a closed-loop interactive process, subsequent observations can correct some geometric deviations caused by single-step depth noise.

\subsection{Anchor-Trajectory Memory}
\label{subsec:memory}

Traditional VLN memory mechanisms often face a dilemma between ``full storage leading to context overflow'' and ``sparse sampling leading to information loss.'' We argue that while non-critical nodes are redundant in high-level semantics, they carry indispensable spatiotemporal transition information. Based on this, the memory is constructed via two components: \textbf{Explicit Anchors} (i.e., the specific decision nodes where CoT reasoning is triggered) and \textbf{Space-Time Indicators}

Designed to inject precise geometric positions ($x, y, yaw$) and temporal ($t$) causality into the context stream, the STI utilizes a feature fusion approach. Specifically, we separately encode spatial coordinates, temporal indices, and angular information using 2D and 1D Rotary Position Embeddings (RoPE) (\cite{su2024roformer}). These embeddings are then concatenated and projected through a Multi-Layer Perceptron (MLP) to form a unified STI token:
\begin{equation}
\label{eq:sti_new}
    E_{STI}(t, x, y, yaw) = \text{MLP} \Big( \big[ \text{RoPE}_{2D}(x, y) ; \text{RoPE}_{1D}(t) ; \text{RoPE}_{2D}(\sin(yaw), \cos(yaw)) \big] \Big)
\end{equation}
where $[ ; ]$ denotes the concatenation operation. Notably, instead of directly encoding the scalar $yaw$, we map it to its continuous trigonometric representation $(\sin(yaw), \cos(yaw))$ to prevent the discontinuity at the $0^\circ/360^\circ$ boundary, ensuring smooth orientation encoding. The MLP projects these concatenated high-frequency trigonometric features into the model's native hidden dimension. The resulting $E_{STI}$ is then added into the input sequence as a single token. This explicitly embeds the absolute spatio-temporal state into a fixed-length space, providing the VLM with dense geometric and temporal awareness for navigation (detailed analysis in \ref{sec:appendix_STI_token}).

Based on this geometric foundation, the memory is constructed via two heterogeneous components: \textbf{Explicit Anchors} and \textbf{Space-Time Indicators}. When the selective reasoning mechanism triggers a Chain-of-Thought (CoT) at a critical topological node $t_k$, the state is stored as a triplet:
\begin{equation}
    A_k = \langle M_{STI}^{(k)}, M_{vis}^{(k)}, M_{state}^{(k)} \rangle
\end{equation}
Here, $M_{STI}^{(k)} = E_{STI}(t_k, x_k, y_k, yaw_k)$ provides the explicit spatiotemporal coordinate; $M_{vis}^{(k)} = v_{t_k}$ retains high-fidelity raw visual features to support pixel-level loop closure detection; and $M_{state}^{(k)}$ acts as a ``Planning Beacon'' by storing CoT to represent the task phase localization, serving as a semantic summary of navigation progress. For the redundant path interval $\mathcal{T}$ between two anchors, the model discards redundant visual features and retains only a pure spatiotemporal stream:
\begin{equation}
    T_k = [E_{STI}(t, x_t, y_t, yaw_t) \mid t \in \mathcal{T}]
\end{equation}
This mechanism allows the memory to maintain full-resolution movement trajectories in non-critical intervals, effectively preventing topological information loss while concentrating expensive visual perception overhead on critical anchors.

At each decision step, the dynamic input context $C_t$ concatenates a Working Memory (comprising two uniformly sampled frames between the current step and the most recent reasoning anchor) with the Long-term Topological Memory (the reconstructed sequence $M_{long} = \{A_1, T_1, \dots, A_m, T_m\}$).

\subsection{Two-Level GRPO Paradigm}
\label{subsec:grpo_rl}

To ensure the model possesses robust priors before the RL phase, we first perform a brief Supervised Fine-Tuning (SFT) (\cite{ouyang2022training}) cold start on the MultiNav-CoT dataset, using Qwen2.5-VL-7B(\cite{Qwen2.5-VL}) as base model. After this cold start, we employ \textbf{Group Relative Policy Optimization (GRPO)} (\cite{shao2024deepseekmath}) for further fine-tuning. We designed a \textbf{Dual Reward System}, combined with group-based relative optimization, to simultaneously enhance fine-grained operational skills and long-range planning capabilities.

\textbf{3.3.1 Local Step Rewards}

Local step rewards $R_{local}^{(t)}$ guide the agent to make logical, safe, and timely reasoning decisions at each step. We define five explicit components: \textbf{Target Approach Reward ($r_{app}^{(t)}$)} encourages continuous movement toward the target; \textbf{Collision Avoidance Reward ($r_{coll}^{(t)}$)} ensures physical traversability; \textbf{Stop Action Reward ($r_{stop}^{(t)}$)} provides supervision for termination;\textbf{Reasoning Value Reward ($r_{rea}^{(t)}$)} evaluates the actual utility of reasoning by correlating CoT indicators with approach scores; 
and \textbf{Format Adherence Reward ($r_{fmt}^{(t)}$)} ensures valid spatial constraints and JSON formatting. The Total Local Reward is defined as (advantage calculations and hyperparameter are in Appendix \ref{app:rewards}):
\begin{equation}
    R_{local}^{(t)} = \lambda_1 r_{app}^{(t)} + \lambda_2 r_{coll}^{(t)} + \lambda_3 r_{stop}^{(t)} + \lambda_4 r_{rea}^{(t)} + \lambda_5 r_{fmt}^{(t)}
\end{equation}

\textbf{3.3.2 Global Trajectory Rewards}

The global reward evaluates the holistic quality of the completed trajectory, aggregating three metrics: \textbf{Task Success Reward ($r_{suc}$)} is the primary driver relaxing strict success to oracle boundaries; \textbf{Trajectory Efficiency Reward ($r_{spl}$)} optimizes navigation efficiency via SPL (\cite{anderson2018evaluation}); and \textbf{Reasoning Density Reward ($r_{den}$)} provides a non-negative efficiency bonus that discourages excessive CoT generation. The Total Global Reward is defined as(advantage calculations and hyperparameter are in Appendix \ref{app:rewards}):
\begin{equation}
    R_{global} = \omega_1 r_{suc} + \omega_2 r_{spl} + \omega_3 r_{den}
\end{equation}
This reward design encourages the agent to minimize cognitive overhead while ensuring task success, naturally guiding it to identify critical decision nodes and eliminate redundant reasoning.

\textbf{3.3.3 Two-Level Policy Optimization}

\begin{figure}[t]
    \centering
\includegraphics[width=\textwidth]{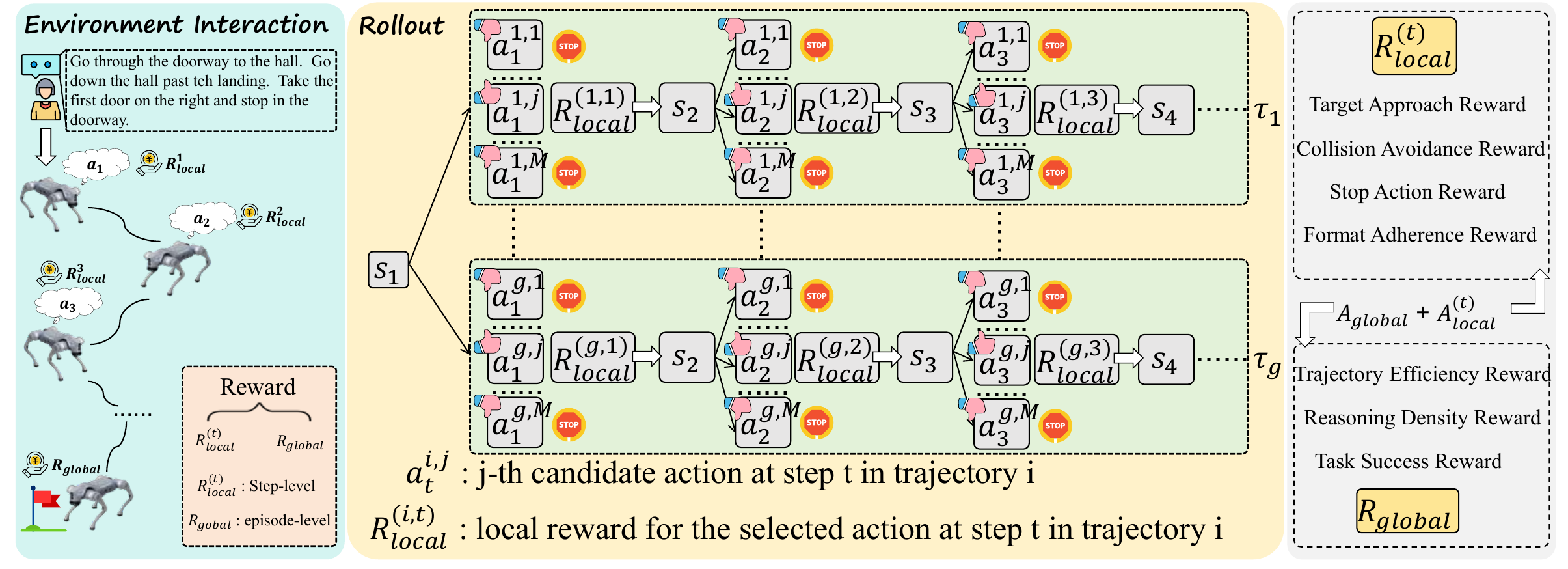}
    \caption{The Two-Level GRPO paradigm. TAMP-Nav superimposes trajectory-level rollouts (competing for global navigation success) with step-level candidate rollouts (exploring diverse 2D visual-spatial actions via VLM temperature sampling).}
    \label{fig:two_level_grpo}
\end{figure}

Unlike previous methods that rely on step-by-step expert trajectories, our GRPO paradigm (Figure \ref{fig:two_level_grpo}) encourages autonomous exploration by superimposing advantages from two hierarchical levels of rollouts. For a given instruction $q$, we generate a group of $G=8$ complete navigation trajectories. To construct each trajectory, at each decision step $t$, the policy generates a candidate set $\mathcal{A}_t = \{a_1, a_2, \dots, a_M\}$ containing $M=4$ actions via temperature-based sampling based on the current context $C_t$. We then evaluate these candidates using the local step reward $R_{local}^{(t)}$. 

To ensure a healthy variance of trajectory qualities for effective gradient updates, we introduce an annealed guided sampling strategy during early RL stages. Specifically, the probability of actually selecting candidate $a_i \in \mathcal{A}_t$ to execute and continue the trajectory is formulated as:
\begin{equation}
    P_{select}(a_i) = \frac{\exp \left( \beta_k \cdot R_{local}^{(t)}(a_i) \right)}{\sum_{j=1}^M \exp \left( \beta_k \cdot R_{local}^{(t)}(a_j) \right)}
\end{equation}
where $\beta_k$ is a guidance coefficient at the global RL training iteration $k$, explicitly defined as an exponential decay schedule $\beta_k = \beta_0 \cdot \alpha^k$ with initial value $\beta_0 > 0$ and decay rate $\alpha \in (0, 1)$. During early stages ($\beta_k > 0$), this explicitly weights the probability of selecting a candidate by its local reward. As training progresses and $\beta_k \to 0$, $P_{select}(a_i)$ approaches a uniform distribution $1/M$. Once the $G$ trajectories are generated, we compute $A_{global}$ and $A_{local}$ via separate standardizations. Specifically, $A_{global}$ is obtained by Z-score standardizing the global trajectory rewards across the $G$ rollouts. Concurrently, at each decision step, the local advantage $A_{local}^{(t)}$ for the actually executed action $a_i$ is derived by Z-score standardizing its local reward $R_{local}^{(t)}(a_i)$ against the $M$ candidates in $\mathcal{A}_t$. By independently normalizing both advantages to a standard normal distribution ($\mathcal{N}(0, 1)$), we align their statistical scales to facilitate a more balanced integration into the superimposed total advantage $A_S^{(t)}$:
\begin{equation}
    A_S^{(t)} = A_{global} + A_{local}^{(t)}
\end{equation}
The policy is updated via the standard GRPO loss formulas provided in Appendix \ref{app:rewards}.

\section{Experiments}
\label{sec:experiments}

\subsection{Experimental Setup}

\textbf{Evaluation Strategy and Metrics.} To rigorously assess TAMP-Nav, we conduct comprehensive experiments across three dimensions: (1) Standard benchmarking on VLN-CE datasets;
(2) In-depth analytical experiments on emergent cognitive abilities and training efficiency; and (3) Ablation studies on key architectural components.
We report performance using standard metrics: Success Rate (SR), Oracle Success Rate (OSR), Success weighted by Path Length (SPL), normalized Dynamic Time Warping (nDTW), and Navigation Error (NE).

\textbf{Implementation Details.} 
Both simulation and real-world experiments utilize a context window of 4,096 tokens. Training, evaluation, and real-world deployment were all conducted on NVIDIA A800 GPUs. Detailed training configurations and hyperparameters are provided in Appendix \ref{sec:appendix_training_configs}.
\subsection{Main Results}

\textbf{Performance on VLN-CE Benchmarks.}
As shown in Table \ref{tab:method-comparison}, TAMP-Nav achieves State-of-the-Art (SOTA) performance across all metrics on the R2R-CE and RxR-CE validation unseen splits (66.2\% on R2R-CE Val-Unseen and 65.7\% on RxR-CE Val-Unseen). A representative qualitative example of the agent's execution in the simulation environment is provided in Appendix \ref{sec:appendix_sim_example}.

Beyond navigation accuracy, TAMP-Nav demonstrates high data and computational efficiency. It achieves competitive results using only 90k training trajectories (700k interactions), significantly fewer than DualVLN (763k trajectories) and NavFoM (3.37M interactions). We attribute this sample efficiency to our dense-reward RL optimization, which improves the success rate on R2R-CE from 55.7\% (SFT-only) to 66.2\%, proving more effective than pure supervised learning in internalizing complex logic from limited data. Computationally, the Pixel-to-3D action paradigm reduces average interaction steps to 9 per trajectory, whereas open-source state-of-the-art models such as DualVLN and StreamVLN require approximately 30 steps. Benefiting from this compacted interaction horizon, sparse reasoning strategy, and the lightweight memory mechanism, TAMP-Nav achieves highly efficient inference: it averages 16.58s per task on a single A800 GPU, significantly faster than StreamVLN (37.47s) and DualVLN (41.46s).

\begin{table}[t]
  \caption{Comparison of different methods on R2R-CE and RxR-CE validation unseen splits.}
  \label{tab:method-comparison}
  \centering
  \resizebox{\textwidth}{!}{
  \begin{tabular}{l cccc cccc}
    \toprule
    & \multicolumn{4}{c}{R2R-CE Val-Unseen} & \multicolumn{4}{c}{RxR-CE Val-Unseen} \\
    \cmidrule(r){2-5} \cmidrule(l){6-9}
    Method & NE$\downarrow$ & OS$\uparrow$ & SR$\uparrow$ & SPL$\uparrow$ & NE$\downarrow$ & SR$\uparrow$ & SPL$\uparrow$ & nDTW$\uparrow$ \\
    \midrule
    HPN+DN* \cite{krantz2021waypoint} & 6.31 & 40.0 & 36.0 & 34.0 & - & - & - & - \\
    CMA* \cite{hong2022bridging} & 6.20 & 52.0 & 41.0 & 36.0 & 8.76 & 26.5 & 22.1 & 47.0 \\
    GridMM* \cite{wang2023gridmm} & 5.11 & 61.0 & 49.0 & 41.0 & - & - & - & - \\
    ETPNav* \cite{an2024etpnav} & 4.71 & 65.0 & 57.0 & 49.0 & 5.64 & 54.7 & 44.8 & 61.9 \\
    ScaleVLN* \cite{wang2023scaling} & 4.80 & - & 55.0 & 51.0 & - & - & - & - \\
    \midrule
    InstructNav \cite{long2024instructnav} & 6.89 & - & 31.0 & 24.0 & - & - & - & - \\
    R2R-CMTP \cite{chen2021topological} & 7.90 & 38.0 & 26.4 & 22.7 & - & - & - & - \\
    LAW \cite{raychaudhuri2021law} & 6.83 & 44.0 & 35.0 & 31.0 & 10.90 & 8.0 & 8.0 & 38.0 \\
    CM2 \cite{georgakis2022cross} & 7.02 & 41.5 & 34.3 & 27.6 & - & - & - & - \\
    WS-MGMap \cite{chen2022weakly} & 6.28 & 47.6 & 38.9 & 34.3 & - & - & - & - \\
    ETPNav + FF \cite{wang2024sim} & 5.95 & 55.8 & 44.9 & 30.4 & 8.79 & 25.5 & 18.1 & - \\
    Seq2Seq \cite{krantz_vlnce_2020} & 7.77 & 37.0 & 25.0 & 22.0 & 12.10 & 13.9 & 11.9 & 30.8 \\
    CMA \cite{krantz_vlnce_2020} & 7.37 & 40.0 & 32.0 & 30.0 & - & - & - & - \\
    \midrule
    VLN-R1 \cite{qi2025vln} & 7.0 & 41.2 & 30.2 & 21.8 & 9.1 & 22.7 & 17.6 & - \\
    NaVid \cite{zhang2024navid} & 5.47 & 49.1 & 37.4 & 35.9 & - & - & - & - \\
    MapNav \cite{zhang2025mapnav} & 4.93 & 53.0 & 39.7 & 37.2 & - & - & - & - \\
    NaVILA \cite{cheng2024navila} & 5.22 & 62.5 & 54.0 & 49.0 & 6.77 & 49.3 & 44.0 & 58.8 \\
    UniNaVid \cite{zhang2024uni} & 5.58 & 53.3 & 47.0 & 42.7 & 6.24 & 48.7 & 40.9 & - \\
    StreamVLN \cite{wei2025streamvln} & 4.98 & 64.2 & 56.9 & 51.9 & 6.22 & 52.9 & 46.0 & 61.9 \\
    NavFoM \cite{zhang2025embodied} & 4.61 & 72.1 & 61.7 & 55.3 & 4.74 & 64.4 & 56.2 & 65.8 \\
    DualVLN \cite{wei2025ground} & 4.05 & 70.7 & 64.3 & 58.5 & 4.58 & 61.4 & 51.8 & 70.0 \\
    TAMP-Nav(only SFT) & 4.88 & 62.0 & 55.7 & 50.3 & 6.10 & 52.4 & 46.2 & 62.1 \\
    \textbf{TAMP-Nav} & \textbf{3.85} & \textbf{74.5} & \textbf{66.2} & \textbf{58.8} & \textbf{4.32} & \textbf{65.7} & \textbf{56.9} & \textbf{72.4} \\
    \bottomrule
  \end{tabular}
  }
\end{table}
\textbf{Performance on Long-Horizon Tasks.}
To further investigate TAMP-Nav's performance stability over extended distances, we evaluate TAMP-Nav on a subset of long-horizon trajectories from the R2R-CE and RxR-CE val-unseen splits. This subset comprises 5927 trajectories where the expert path length exceeds 12.5 meters (equivalent to over 50 atomic forward actions). As shown in Table \ref{tab:ablation_memory}, TAMP-Nav (AT-Mem) significantly outperforms strong baselines, achieving a Success Rate (SR) of 49.8\%, which is a substantial improvement over StreamVLN (30.9\%) and DualVLN (41.9\%). Figure \ref{fig:long_horizon_dist} further breaks down performance across varying trajectory lengths.

\subsection{Emergent Reasoning-on-Demand}

During the inference phase, TAMP-Nav autonomously determines \textit{when} to trigger selective reasoning. To verify this capability, we analyze the spatial distribution of CoT triggers alongside comparative intervention experiments across different trigger strategies.

\begin{figure}[h]
    \centering
    \begin{minipage}[t]{0.58\textwidth}
        \vspace{0pt}
        \centering
        \includegraphics[width=\textwidth]{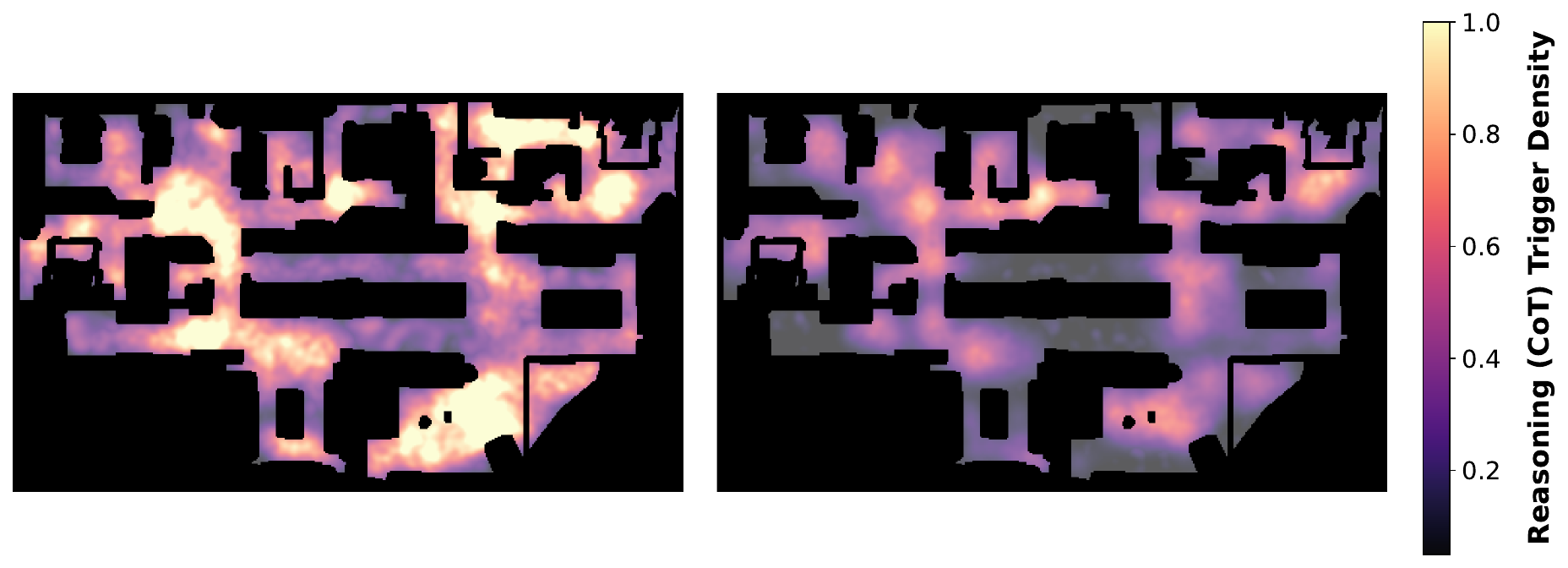}
        \captionof{figure}{\textbf{Spatial Heatmap of Reasoning (CoT) Triggers.} Comparison between the SFT model (left) and the RL-aligned TAMP-Nav (right). }
        \label{fig:heatmap}
    \end{minipage}
    \hfill
    \begin{minipage}[t]{0.38\textwidth}
        \vspace{0pt}
        \centering
        \captionof{table}{\textbf{Intervention Analysis.} Comparing navigation success (SR) and reasoning density (CoT Ratio) on R2R-CE Val-Unseen.}
        \label{tab:intervention}
        \small
        \begin{tabular}{l c c}
            \toprule
            Strategy & CoT Ratio & SR \\
            \midrule
            Dense CoT & 100\% & 66.8\% \\
            Fixed (1/3) & 36.2\% & 60.1\% \\
\midrule
            \textbf{Auto (Ours)} & \textbf{26.3\%} & \textbf{66.2\%} \\
            \bottomrule
        \end{tabular}
    \end{minipage}
\end{figure}

As illustrated in Figure \ref{fig:heatmap}, we recorded the locations where the model generated \texttt{<think>} tokens and applied Gaussian Kernel Density Estimation (KDE) to visualize their spatial distribution. The heatmaps indicate that the RL-trained TAMP-Nav concentrates reasoning at critical topological nodes, such as crossroads, doorways, and near target objects. Quantitatively, the proportion of reasoning steps allocated to straight corridors decreases from 38\% in the SFT model to 11\% after RL alignment. This indicates that the agent learns to suppress redundant CoT in trivial intervals, prioritizing computation for complex decision junctures.

Table \ref{tab:intervention} evaluates this emergent timing. Remarkably, TAMP-Nav nearly matches the \textit{Dense CoT} upper bound (66.8\% vs. 66.2\% SR) with only a 26.3\% CoT ratio, significantly outperforming the \textit{Fixed-Interval} baseline. Supported by Figure \ref{fig:heatmap}, these results confirm our RL formulation effectively incentivizes the agent to bypass redundant thinking, triggering reasoning at optimal moments.
\subsection{Robustness to Depth Uncertainty}
\label{subsec:depth_robustness}

To assess robustness against sensor inaccuracies, we inject multiplicative Gaussian noise into the depth of the selected pixel:
\begin{equation}
    \tilde{d}_t = D_t(u, v) \cdot \epsilon, \quad \epsilon \sim \mathcal{N}(1, \sigma^2)
\end{equation}
where $\sigma$ is the relative noise intensity. To mimic low-level collision avoidance, any perturbed waypoint in non-navigable regions is snapped to the nearest valid point.

Under the tested multiplicative-noise protocol, SR decreases by 2.8 points at $\sigma=0.2$ (16\% relative error), indicating robustness to moderate waypoint-depth perturbations.

\begin{table}[h]
  \centering
\begin{minipage}[c]{0.48\textwidth} \centering
    \caption{Success Rate (SR) under varying multiplicative depth noise levels ($\sigma$).}
    \label{tab:depth_noise}
    \vspace{2pt}
    \footnotesize
    \setlength{\tabcolsep}{4pt}
    \begin{tabular}{lcccc}
        \toprule
        Noise $\sigma$ & 0 & 0.05 & 0.1 & 0.2 \\
        \midrule
        SR (\%) & \textbf{66.2} & 65.9 & 65.3 & 63.4 \\
        $\Delta$ SR & - & -0.3 & -0.9 & -2.8 \\
        \bottomrule
    \end{tabular}
  \end{minipage}
  \hfill 
\begin{minipage}[c]{0.48\textwidth} \centering
    \caption{Performance on long-horizon tasks and memory ablation study.}
    \label{tab:ablation_memory}
    \vspace{2pt}
    \footnotesize
    \setlength{\tabcolsep}{6pt}
    \begin{tabular}{lc} 
      \toprule
      Model & SR ($\uparrow$) \\
      \midrule
      StreamVLN & 30.9 \\
      DualVLN & 41.9 \\
      TAMP-Nav(US) & 40.5 \\ 
      TAMP-Nav(Full) & 42.4 \\ 
      TAMP-Nav(w/o STI) & 45.6 \\
      \textbf{TAMP-Nav(AT-Mem)} & \textbf{49.8} \\
      \bottomrule
    \end{tabular}
  \end{minipage}
\end{table}
\subsection{Ablation Studies}

\textbf{4.5.1 Controlled Component Analysis.}\par

We conduct block-wise ablations on the R2R-CE and RxR-CE val-unseen splits. All variants are initialized from Qwen2.5-VL-7B and use the same sensing inputs, data split, fixed non-learned SLAM controller, and evaluation protocol. Each block varies one component while holding the remaining configuration fixed; comparisons are therefore made within each block.

\begin{table}[htbp]
  \color{black}
  \centering
  \caption{\textbf{Controlled component-wise ablations under matched budgets.} R2R-CE reports NE/OS/SR/SPL; RxR-CE reports NE/SR/SPL/nDTW.}
  \label{tab:controlled_ablation}
  \small
  \setlength{\tabcolsep}{4pt}
  \resizebox{\textwidth}{!}{
  \begin{tabular}{llcc}
    \toprule
    Block & Variant & R2R-CE: NE / OS / SR / SPL & RxR-CE: NE / SR / SPL / nDTW \\
    \midrule
    Action (SFT only) & NavFoM-style metric waypoint & 7.18 / 42.0 / 30.9 / 24.7 & 8.47 / 27.2 / 20.6 / 43.1 \\
     & Pixel-to-3D & 4.88 / 62.0 / 55.7 / 50.3 & 6.10 / 52.4 / 46.2 / 62.1 \\
    \midrule
    GRPO & Global trajectory advantage only & 4.52 / 67.0 / 59.3 / 52.1 & 5.30 / 58.4 / 49.7 / 65.5 \\
     & Two-Level without annealed guidance & 4.12 / 71.5 / 63.4 / 56.0 & 4.72 / 62.6 / 53.8 / 69.4 \\
     & Full Two-Level GRPO & 3.85 / 74.5 / 66.2 / 58.8 & 4.32 / 65.7 / 56.9 / 72.4 \\
    \midrule
    Memory & Uniform sampling & 4.43 / 68.2 / 60.8 / 53.6 & 5.03 / 59.4 / 50.8 / 66.8 \\
     & Full history & 4.29 / 69.4 / 61.9 / 54.7 & 4.87 / 61.1 / 51.8 / 68.0 \\
     & AT-Mem without STI & 4.08 / 71.4 / 63.6 / 56.1 & 4.59 / 62.4 / 52.7 / 69.8 \\
     & Full AT-Mem & 3.85 / 74.5 / 66.2 / 58.8 & 4.32 / 65.7 / 56.9 / 72.4 \\
    \midrule
    Reasoning & Dense CoT & 3.79 / 75.1 / 66.8 / 59.3 & 4.25 / 66.1 / 57.4 / 72.8 \\
     & Fixed (1/3) & 4.46 / 67.6 / 60.1 / 52.8 & 5.10 / 58.9 / 50.0 / 65.9 \\
     & Auto & 3.85 / 74.5 / 66.2 / 58.8 & 4.32 / 65.7 / 56.9 / 72.4 \\
    \bottomrule
  \end{tabular}}
\end{table}

The SFT-only action block uses 90k trajectories, approximately 700k interactions, and one training epoch without GRPO. Replacing Pixel-to-3D with a NavFoM-style metric-waypoint output while keeping the remaining architecture fixed reduces SR by 24.8/25.2 points on R2R-CE/RxR-CE. In the GRPO block, local advantages improve SR by 4.1/4.2 points over trajectory-level advantage alone, and annealed guidance contributes a further 2.8/3.1 points. Full AT-Mem improves over full-history retention by 4.3/4.6 points, while removing STI decreases SR by 2.6/3.3 points. Auto reduces R2R-CE CoT calls by 73.7\% relative to Dense CoT with a 0.6-point SR decrease; the corresponding SR decrease on RxR-CE is 0.4 points.

\textbf{4.5.2 Influence of CoT Supervision.}\par

We evaluate the effect of the CoT supervision source under a matched SFT budget. All settings share the same action labels, training trajectories, navigation policy architecture, and one-epoch schedule; only the reasoning annotations differ.

\begin{table}[htbp]
  \color{black}
  \centering
  \caption{\textbf{Effect of the CoT supervision source under a matched SFT budget.}}
  \label{tab:cot_teacher_ablation}
  \small
  \setlength{\tabcolsep}{4pt}
  \resizebox{\textwidth}{!}{
  \begin{tabular}{lcc}
    \toprule
    SFT supervision & R2R-CE: NE / OS / SR / SPL & RxR-CE: NE / SR / SPL / nDTW \\
    \midrule
    Action only (no CoT) & 5.31 / 57.0 / 50.6 / 45.0 & 6.71 / 46.1 / 39.8 / 56.2 \\
    Qwen2.5-VL-7B CoT & 5.46 / 55.7 / 49.4 / 43.1 & 6.83 / 44.8 / 38.2 / 54.9 \\
    GLM-4.5V-108B CoT & 4.99 / 60.8 / 54.4 / 48.9 & 6.24 / 51.2 / 44.9 / 60.8 \\
    Gemini 2.5 Flash CoT & 4.88 / 62.0 / 55.7 / 50.3 & 6.10 / 52.4 / 46.2 / 62.1 \\
    \bottomrule
  \end{tabular}}
\end{table}

Qwen2.5-VL-7B supervision performs below the Action-only setting on both benchmarks. Inspection of its annotations reveals frequent spatial contradictions, incorrect task-phase identification, and inconsistencies between the rationale and waypoint. GLM-4.5V-108B substantially improves over the smaller teacher and remains within 1.3/1.2 SR points of Gemini 2.5 Flash on R2R-CE/RxR-CE. We further conduct a human quality assessment in which two annotators independently evaluate 200 randomly sampled CoTs from each of the two stronger teachers for instruction phase, visual--spatial consistency, rationale--waypoint consistency, and completeness. Gemini and GLM achieve pass rates of 94.0\% and 91.5\%, respectively. These results show that downstream navigation performance is sensitive to the spatial consistency of CoT supervision, while the gap between the strong open-weight and proprietary teachers is modest.

\textbf{4.5.3 Effectiveness of Memory Mechanisms.}\par
As shown in Table \ref{tab:ablation_memory}, we compare our Anchor-Trajectory Memory with traditional history management strategies and ablate its core component. \textbf{TAMP-Nav(US)} and \textbf{TAMP-Nav(Full)} represent conventional memory designs relying solely on raw visual features via uniform sampling or full history retention. AT-Mem effectively overcomes the critical information loss inherent in Uniform Sampling and the severe attention dilution suffered by Full History. Furthermore, removing \textbf{Space-Time Indicators (w/o STI)} while preserving other AT-Mem components drops performance from 49.8\% to 45.6\% SR. This demonstrates that while visual anchors provide semantic grounding, lightweight STI tokens are essential for maintaining long-horizon spatial memory.

\begin{figure}[htbp]
    \centering
    \includegraphics[width=\textwidth]{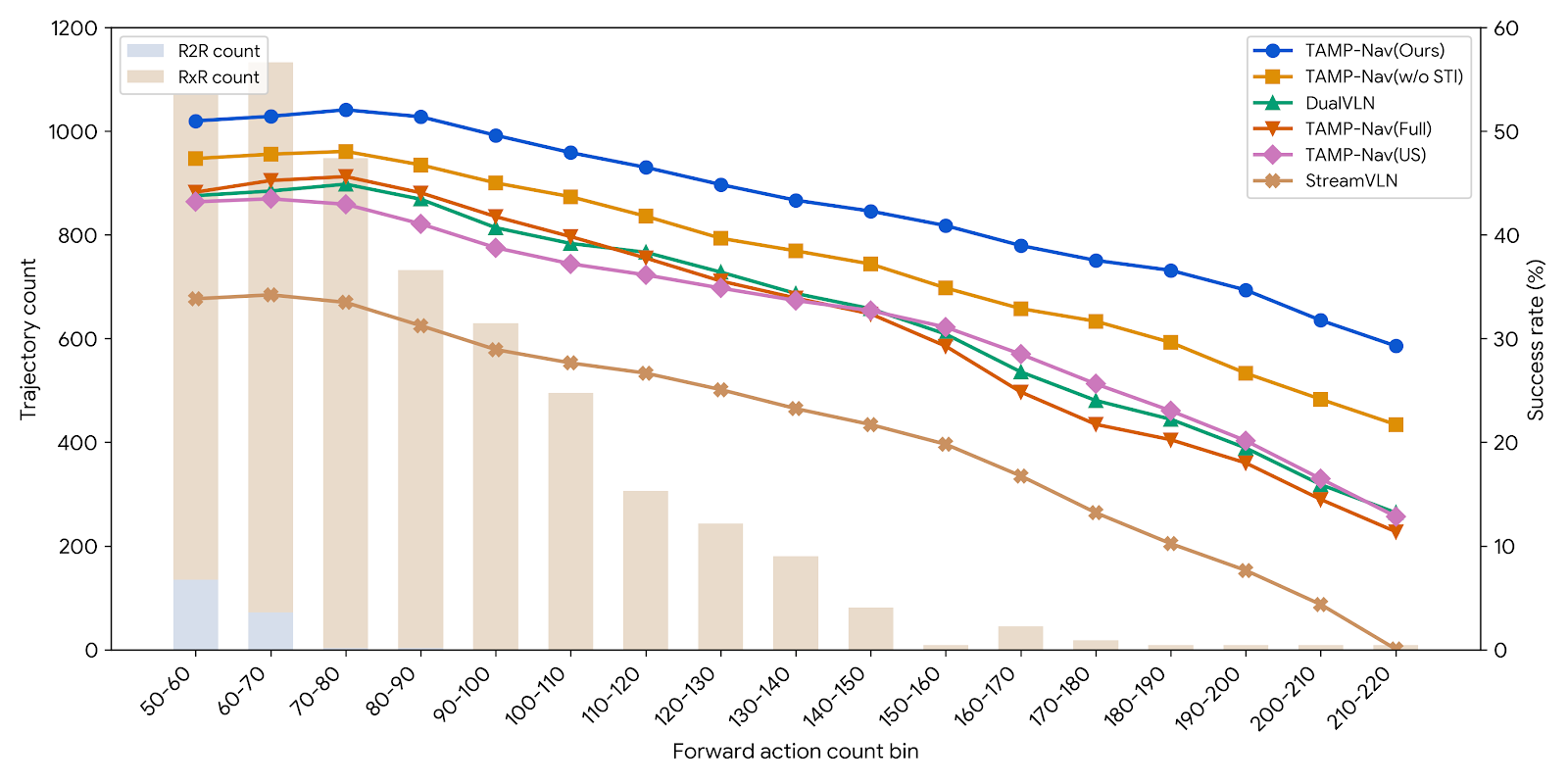}
    \caption{\textbf{Performance distribution on long-horizon navigation tasks.} The bars show the trajectory count in each length bin, and the curves show the Success Rate (SR) of the evaluated methods as trajectory length increases.}
    \label{fig:long_horizon_dist}
\end{figure}

To examine how these memory mechanisms behave as trajectories grow longer, Figure \ref{fig:long_horizon_dist} groups the combined R2R-CE and RxR-CE long-horizon subset by expert trajectory length. Each atomic forward action corresponds to a 0.25-meter movement; bins span from 50 to more than 210 actions in intervals of 10. The bars show the number of trajectories in each bin, while the curves report the SR of each method.

Three trends emerge. First, TAMP-Nav maintains a performance margin over StreamVLN and DualVLN across all length bins; for example, in the 150--160-action bin, TAMP-Nav achieves approximately 40\% SR, whereas StreamVLN drops to approximately 20\%. Second, under the same fixed token budget, Uniform Sampling spans the trajectory using a budget-dependent sampling interval, whereas Full History retains the most recent frames after the budget is exceeded. Uniform Sampling increasingly loses critical observations as paths lengthen, while Full History suffers from attention dilution. In contrast, AT-Mem preserves critical visual anchors and compresses intermediate motion into lightweight trajectory tokens. Third, the gap between full TAMP-Nav and \textbf{TAMP-Nav (w/o STI)} widens on longer trajectories (e.g., beyond 120 actions), indicating that STI tokens help preserve geometric connectivity over extended execution.

\begin{figure}[h]
    \centering
    \includegraphics[width=0.65\textwidth]{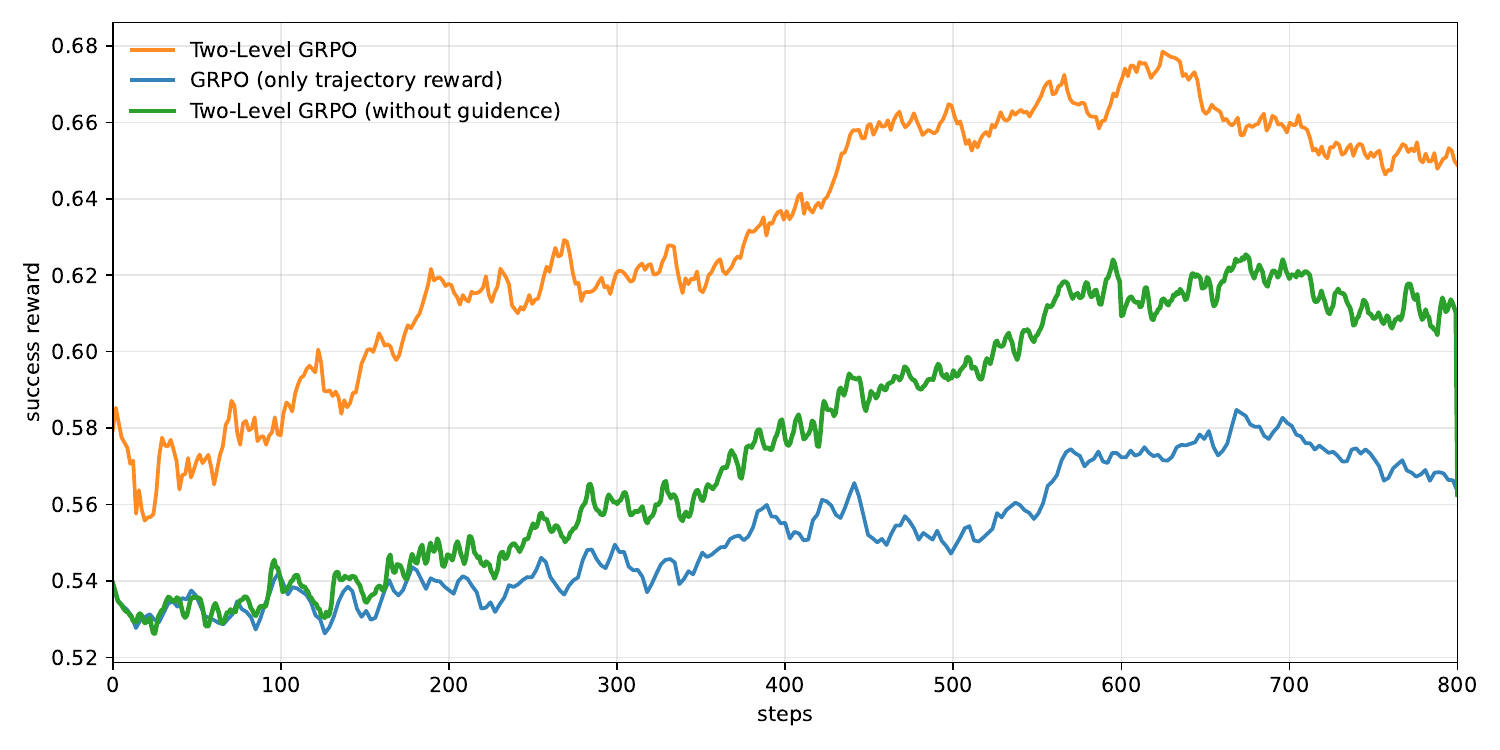}
    \caption{\textbf{Learning curves of success rewards under different GRPO configurations.} The plot illustrates the training progress over 800 steps for three variants: the full Two-Level GRPO, the version without annealed guided sampling (``without guidance''), and the standard GRPO using only trajectory-level rewards (``only trajectory reward'').}
    \label{fig:grpo_dynamics}
\end{figure}

\textbf{4.5.4 Effectiveness of the Two-Level GRPO Framework}\par

To evaluate our RL design, we analyze the learning dynamics of TAMP-Nav under different GRPO configurations. Figure \ref{fig:grpo_dynamics} compares the success reward across 800 training steps for standard GRPO (trajectory rewards only), Two-Level GRPO without guidance, and the full framework.

\textbf{Overcoming Sparse Feedback.} The baseline model trained solely with trajectory-level rewards exhibits training instability. In long-horizon navigation, the delay of terminal rewards makes it difficult for the model to associate final outcomes with specific intermediate reasoning or spatial decisions. This results in slower learning and convergence at a 0.59 reward. By superimposing local step advantages, our Two-Level GRPO provides immediate signals that stabilize intermediate state optimization, lifting the reward to 0.64 even without guided sampling.

\textbf{Impact of Guided Sampling.} The full Two-Level GRPO, which incorporates annealed guided sampling, achieves faster convergence and a higher final performance (0.68). This indicates that in early RL stages, purely random temperature sampling can produce rollouts with limited informative value. By weighting candidate selection based on local step rewards, guided sampling increases the density of high-quality trajectories in early samples, thereby accelerating policy updates and improving exploration efficiency.

\subsection{Real-World Deployment}

\begin{figure}[htbp]
    \centering
\begin{subfigure}[b]{0.65\textwidth}
        \centering
        \includegraphics[width=\textwidth]{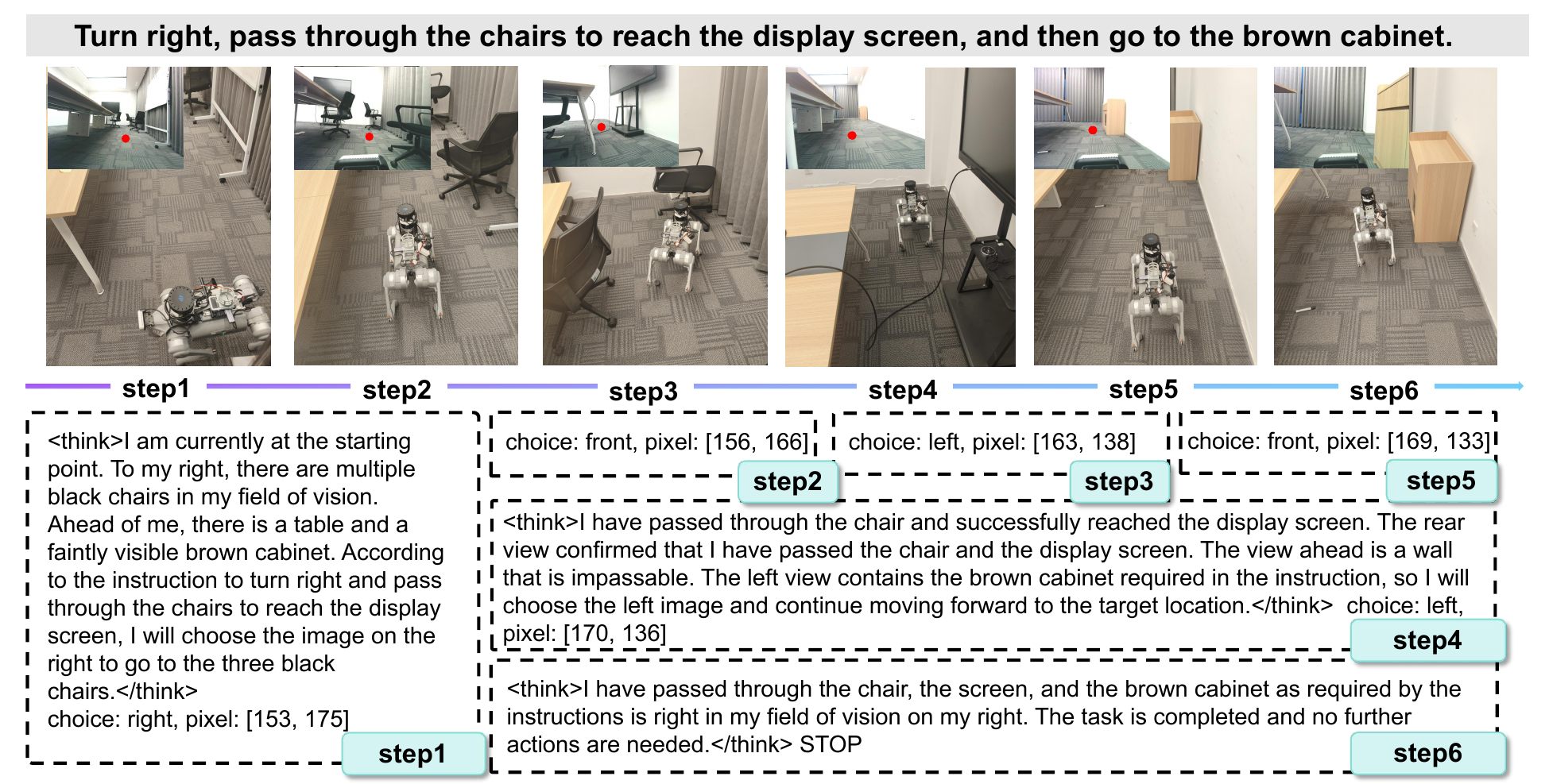}
        \caption{Visualization of real-world experiments}
        \label{fig:real_world_vis}
    \end{subfigure}
    \hfill
\begin{subfigure}[b]{0.30\textwidth}
        \centering
        \includegraphics[width=\textwidth]{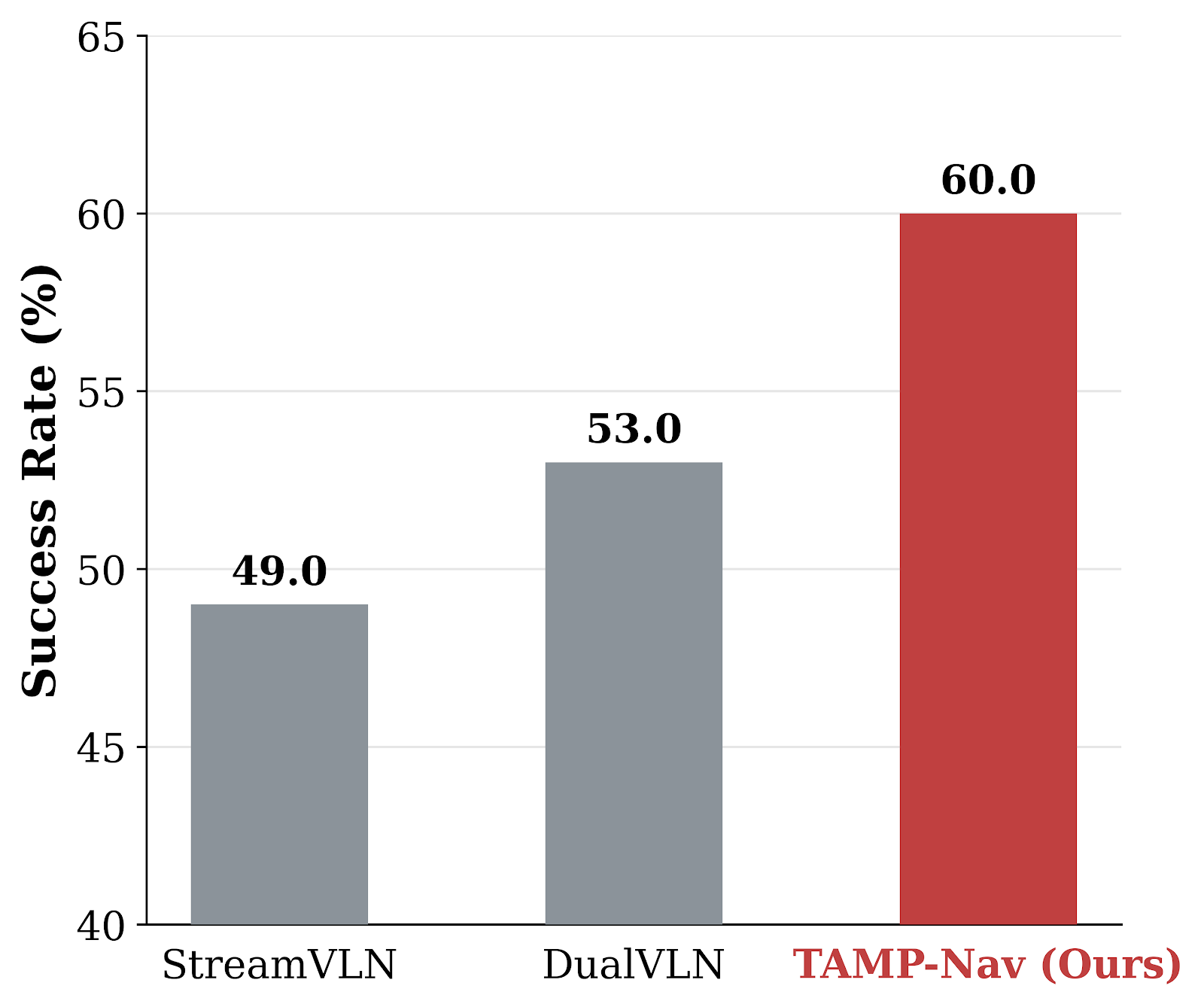}
        \caption{Success rate of real-world experiments}
        \label{fig:real_world_sr}
    \end{subfigure}
    \caption{Real-world deployment results. (a) Visualization of the execution trajectory. (b) Success rate comparison.}
    \label{fig:real_world}
\end{figure}

\textbf{Evaluation protocol.} The evaluation covers five task categories: meeting room, laboratory, hall, cross-area, and outdoor navigation. Each category contains four fixed natural-language instructions, and each instruction is repeated five times. Each method is therefore evaluated in 100 valid trials over the same 20 instructions, start positions, and goal positions, including 80 indoor and 20 outdoor trials. The complete evaluation contains 300 valid trials across the three methods. The surveyed feasible reference routes range from 8.88 to 44.74 meters, with a mean length of 24.55 meters.

\textbf{Outcome and trial-validity criteria.} Each high-level model request is limited to 5 seconds, and the low-level system is allowed 30 seconds to execute each high-level action. A trial is considered successful only if the robot autonomously stops within 1.5 meters of the surveyed target without human navigation or control intervention. Every trial that enters the navigation process normally is included in the evaluation. A trial is counted as a failure if it does not meet the success criterion, exceeds either time limit, or requires the safety operator to activate the hardware emergency stop or assume control because of an imminent collision, fall, or other safety risk. A trial is declared invalid and repeated under the same setting only when the execution logs confirm a robot mechanical fault or experimental equipment fault unrelated to the navigation execution chain. Any problem arising after a normal start from visual perception, model decision-making, localization, planning, control, or stopping is retained as a failure and is neither excluded nor retried.

\textbf{Comparison scope.} All methods share the robot platform, tasks, start and goal positions, time limits, and success criterion, while retaining their recommended sensing and low-level control mechanisms. The physical results therefore compare the complete deployed systems under their respective configurations.

As shown in Figure \ref{fig:real_world}b, without any real-robot fine-tuning, TAMP-Nav achieved a 60.0\% Success Rate, outperforming StreamVLN (49.0\%) and DualVLN (53.0\%). 
Furthermore, Figure \ref{fig:real_world}a illustrates a representative execution where the agent effectively follows instructions and exhibits reasoning-on-demand behavior. Additional trajectory examples are provided in Appendix \ref{sec:appendix_real_world}.

\section{Limitations}
\label{sec:limitations}

TAMP-Nav depends on depth information for Pixel-to-3D projection, STI encoding, and coordinate transformations, and on a reliable SLAM stack for low-level execution. Although the controlled depth-noise study shows only a 2.8-point SR decrease under the strongest tested perturbation, severe depth errors, accumulated odometry or SLAM drift, localization failure, and highly dynamic environments can cause larger degradation.

GRPO training relies on privileged simulator signals for progress, success, SPL, and collision, and the current system does not support online reinforcement learning on a physical robot. Real-world navigation rollouts are substantially slower than simulation, while computing the complete reward online would require endpoint annotations, human feedback, or a reliable learned verifier. Physical deployment is therefore limited to zero-shot transfer of the frozen policy learned in simulation.

The key-node mining heuristic selects CoT supervision using semantic similarity and visual change. It may under-select critical states when decision points are visually non-salient or when instruction styles differ substantially from those in R2R-CE and RxR-CE. Such settings may require additional training examples or a learned triggering criterion.

The current STI token encodes planar position $(x,y)$ and orientation but does not include height and therefore cannot explicitly distinguish floors. Task-progress descriptions retained in memory anchors provide partial compensation but do not replace explicit 3D localization. Extending STI to a full 3D pose representation may improve robustness in more complex multi-floor buildings and uneven terrain. In addition, the reasoning-density reward uses fixed CoT-ratio thresholds of 0.4, where the density bonus begins to decay more steeply, and 0.6, where it reaches zero. These thresholds may require retuning when the useful reasoning density differs substantially from that of R2R-CE and RxR-CE.

\section{Discussion and Conclusion}
\label{sec:conclusion}

In this paper, we propose \textbf{TAMP-Nav}, a unified framework that achieves spatial, cognitive, and optimization alignment for embodied navigation. By introducing the \textbf{Pixel-to-3D Action Formulation}, an integrated \textbf{Selective Reasoning and Anchor-Trajectory Memory} mechanism, and a \textbf{Two-Level GRPO} paradigm, we seamlessly bridge high-level visual-language reasoning with low-level physical execution. This approach advances VLM-based navigators while maintaining high sample and inference efficiency. Although currently trained exclusively on VLN-CE, future work will extend this framework to broader embodied tasks and dynamic real-world environments.
\bibliographystyle{assets/acl_natbib}
\bibliography{en}

@article{zhang2024vision,
  title={Vision-and-language navigation today and tomorrow: A survey in the era of foundation models},
  author={Zhang, Yue and Ma, Ziqiao and Li, Jialu and Qiao, Yanyuan and Wang, Zun and Chai, Joyce and Wu, Qi and Bansal, Mohit and Kordjamshidi, Parisa},
  journal={arXiv preprint arXiv:2407.07035},
  year={2024}
}

@inproceedings{krantz_vlnce_2020,
  title={Beyond the nav-graph: Vision-and-language navigation in continuous environments},
  author={Krantz, Jacob and Wijmans, Erik and Majumdar, Arjun and Batra, Dhruv and Lee, Stefan},
  booktitle={European Conference on Computer Vision},
  pages={104--120},
  year={2020},
  organization={Springer}
}

@inproceedings{ku2020room,
  title={Room-across-room: Multilingual vision-and-language navigation with dense spatiotemporal grounding},
  author={Ku, Alexander and Anderson, Peter and Patel, Roma and Ie, Eugene and Baldridge, Jason},
  booktitle={Proceedings of the 2020 Conference on Empirical Methods in Natural Language Processing (EMNLP)},
  pages={4392--4412},
  year={2020}
}

@article{zhang2024navid,
  title={Navid: Video-based vlm plans the next step for vision-and-language navigation},
  author={Zhang, Jiazhao and Wang, Kunyu and Xu, Rongtao and Zhou, Gengze and Hong, Yicong and Fang, Xiaomeng and Wu, Qi and Zhang, Zhizheng and Wang, He},
  journal={arXiv preprint arXiv:2402.15852},
  year={2024}
}

@article{zhang2024uni,
  title={Uni-navid: A video-based vision-language-action model for unifying embodied navigation tasks},
  author={Zhang, Jiazhao and Wang, Kunyu and Wang, Shaoan and Li, Minghan and Liu, Haoran and Wei, Songlin and Wang, Zhongyuan and Zhang, Zhizheng and Wang, He},
  journal={arXiv preprint arXiv:2412.06224},
  year={2024}
}

@article{cheng2024navila,
  title={Navila: Legged robot vision-language-action model for navigation},
  author={Cheng, An-Chieh and Ji, Yandong and Yang, Zhaojing and Gongye, Zaitian and Zou, Xueyan and Kautz, Jan and B{\i}y{\i}k, Erdem and Yin, Hongxu and Liu, Sifei and Wang, Xiaolong},
  journal={arXiv preprint arXiv:2412.04453},
  year={2024}
}

@article{wei2025streamvln,
  title={Streamvln: Streaming vision-and-language navigation via slowfast context modeling},
  author={Wei, Meng and Wan, Chenyang and Yu, Xiqian and Wang, Tai and Yang, Yuqiang and Mao, Xiaohan and Zhu, Chenming and Cai, Wenzhe and Wang, Hanqing and Chen, Yilun and others},
  journal={arXiv preprint arXiv:2507.05240},
  year={2025}
}

@article{wei2025ground,
  title={Ground slow, move fast: A dual-system foundation model for generalizable vision-and-language navigation},
  author={Wei, Meng and Wan, Chenyang and Peng, Jiaqi and Yu, Xiqian and Yang, Yuqiang and Feng, Delin and Cai, Wenzhe and Zhu, Chenming and Wang, Tai and Pang, Jiangmiao and others},
  journal={arXiv preprint arXiv:2512.08186},
  year={2025}
}

@article{zhang2025embodied,
  title={Embodied navigation foundation model},
  author={Zhang, Jiazhao and Li, Anqi and Qi, Yunpeng and Li, Minghan and Liu, Jiahang and Wang, Shaoan and Liu, Haoran and Zhou, Gengze and Wu, Yuze and Li, Xingxing and others},
  journal={arXiv preprint arXiv:2509.12129},
  year={2025}
}

@article{lin2025navcot,
  title={Navcot: Boosting llm-based vision-and-language navigation via learning disentangled reasoning},
  author={Lin, Bingqian and Nie, Yunshuang and Wei, Ziming and Chen, Jiaqi and Ma, Shikui and Han, Jianhua and Xu, Hang and Chang, Xiaojun and Liang, Xiaodan},
  journal={IEEE Transactions on Pattern Analysis and Machine Intelligence},
  year={2025},
  publisher={IEEE}
}

@inproceedings{song2023llmplanner,
  title={Llm-planner: Few-shot grounded planning for embodied agents with large language models},
  author={Song, Chan Hee and Wu, Jiaman and Washington, Clayton and Sadler, Brian M and Chao, Wei-Lun and Su, Yu},
  booktitle={Proceedings of the IEEE/CVF international conference on computer vision},
  pages={2998--3009},
  year={2023}
}

@article{qi2025vln,
  title={Vln-r1: Vision-language navigation via reinforcement fine-tuning},
  author={Qi, Zhangyang and Zhang, Zhixiong and Yu, Yizhou and Wang, Jiaqi and Zhao, Hengshuang},
  journal={arXiv preprint arXiv:2506.17221},
  year={2025}
}

@inproceedings{yang2022far,
  title={Far planner: Fast, attemptable route planner using dynamic visibility update},
  author={Yang, Fan and Cao, Chao and Zhu, Hongbiao and Oh, Jean and Zhang, Ji},
  booktitle={2022 ieee/rsj international conference on intelligent robots and systems (iros)},
  pages={9--16},
  year={2022},
  organization={IEEE}
}

@article{xu2022fastlio2,
  title={Fast-lio2: Fast direct lidar-inertial odometry},
  author={Xu, Wei and Cai, Yixi and He, Dongjiao and Lin, Jiarong and Zhang, Fu},
  journal={IEEE Transactions on Robotics},
  volume={38},
  number={4},
  pages={2053--2073},
  year={2022},
  publisher={IEEE}
}

@inproceedings{radford2021learning,
  title={Learning transferable visual models from natural language supervision},
  author={Radford, Alec and Kim, Jong Wook and Hallacy, Chris and Ramesh, Aditya and Goh, Gabriel and Agarwal, Sandhini and Sastry, Girish and Askell, Amanda and Mishkin, Pamela and Clark, Jack and others},
  booktitle={International conference on machine learning},
  pages={8748--8763},
  year={2021},
  organization={PmLR}
}

@article{dosovitskiy2020image,
  title={An image is worth 16x16 words: Transformers for image recognition at scale},
  author={Dosovitskiy, Alexey and Beyer, Lucas and Kolesnikov, Alexander and Weissenborn, Dirk and Zhai, Xiaohua and Unterthiner, Thomas and Dehghani, Mostafa and Minderer, Matthias and Heigold, Georg and Gelly, Sylvain and others},
  journal={arXiv preprint arXiv:2010.11929},
  year={2020}
}

@article{comanici2025gemini,
  title={Gemini 2.5: Pushing the frontier with advanced reasoning, multimodality, long context, and next generation agentic capabilities},
  author={Comanici, Gheorghe and Bieber, Eric and Schaekermann, Mike and Pasupat, Ice and Sachdeva, Noveen and Dhillon, Inderjit and Blistein, Marcel and Ram, Ori and Zhang, Dan and Rosen, Evan and others},
  journal={arXiv preprint arXiv:2507.06261},
  year={2025}
}

@article{su2024roformer,
  title={Roformer: Enhanced transformer with rotary position embedding},
  author={Su, Jianlin and Ahmed, Murtadha and Lu, Yu and Pan, Shengfeng and Bo, Wen and Liu, Yunfeng},
  journal={Neurocomputing},
  volume={568},
  pages={127063},
  year={2024},
  publisher={Elsevier}
}

@article{ouyang2022training,
  title={Training language models to follow instructions with human feedback},
  author={Ouyang, Long and Wu, Jeffrey and Jiang, Xu and Almeida, Diogo and Wainwright, Carroll and Mishkin, Pamela and Zhang, Chong and Agarwal, Sandhini and Slama, Katarina and Ray, Alex and others},
  journal={Advances in neural information processing systems},
  volume={35},
  pages={27730--27744},
  year={2022}
}

@article{wei2022cot,
  title={Chain-of-thought prompting elicits reasoning in large language models},
  author={Wei, Jason and Wang, Xuezhi and Schuurmans, Dale and Bosma, Maarten and Xia, Fei and Chi, Ed and Le, Quoc V and Zhou, Denny and others},
  journal={Advances in neural information processing systems},
  volume={35},
  pages={24824--24837},
  year={2022}
}

@article{shao2024deepseekmath,
  title={Deepseekmath: Pushing the limits of mathematical reasoning in open language models},
  author={Shao, Zhihong and Wang, Peiyi and Zhu, Qihao and Xu, Runxin and Song, Junxiao and Bi, Xiao and Zhang, Haowei and Zhang, Mingchuan and Li, YK and Wu, Yang and others},
  journal={arXiv preprint arXiv:2402.03300},
  year={2024}
}

@article{anderson2018evaluation,
  title={On evaluation of embodied navigation agents},
  author={Anderson, Peter and Chang, Angel and Chaplot, Devendra Singh and Dosovitskiy, Alexey and Gupta, Saurabh and Koltun, Vladlen and Kosecka, Jana and Malik, Jitendra and Mottaghi, Roozbeh and Savva, Manolis and others},
  journal={arXiv preprint arXiv:1807.06757},
  year={2018}
}

@inproceedings{krantz2021waypoint,
  title={Waypoint models for instruction-guided navigation in continuous environments},
  author={Krantz, Jacob and Gokaslan, Aaron and Batra, Dhruv and Lee, Stefan and Maksymets, Oleksandr},
  booktitle={Proceedings of the IEEE/CVF International Conference on Computer Vision},
  pages={15162--15171},
  year={2021}
}

@inproceedings{hong2022bridging,
  title={Bridging the gap between learning in discrete and continuous environments for vision-and-language navigation},
  author={Hong, Yicong and Wang, Zun and Wu, Qi and Gould, Stephen},
  booktitle={Proceedings of the IEEE/CVF conference on computer vision and pattern recognition},
  pages={15439--15449},
  year={2022}
}

@inproceedings{wang2023gridmm,
  title={Gridmm: Grid memory map for vision-and-language navigation},
  author={Wang, Zihan and Li, Xiangyang and Yang, Jiahao and Liu, Yeqi and Jiang, Shuqiang},
  booktitle={Proceedings of the IEEE/CVF International conference on computer vision},
  pages={15625--15636},
  year={2023}
}

@article{an2024etpnav,
  title={Etpnav: Evolving topological planning for vision-language navigation in continuous environments},
  author={An, Dong and Wang, Hanqing and Wang, Wenguan and Wang, Zun and Huang, Yan and He, Keji and Wang, Liang},
  journal={IEEE Transactions on Pattern Analysis and Machine Intelligence},
  year={2024},
  publisher={IEEE}
}

@inproceedings{wang2023scaling,
  title={Scaling data generation in vision-and-language navigation},
  author={Wang, Zun and Li, Jialu and Hong, Yicong and Wang, Yi and Wu, Qi and Bansal, Mohit and Gould, Stephen and Tan, Hao and Qiao, Yu},
  booktitle={Proceedings of the IEEE/CVF international conference on computer vision},
  pages={12009--12020},
  year={2023}
}

@article{long2024instructnav,
  title={Instructnav: Zero-shot system for generic instruction navigation in unexplored environment},
  author={Long, Yuxing and Cai, Wenzhe and Wang, Hongcheng and Zhan, Guanqi and Dong, Hao},
  journal={arXiv preprint arXiv:2406.04882},
  year={2024}
}

@inproceedings{chen2021topological,
  title={Topological planning with transformers for vision-and-language navigation},
  author={Chen, Kevin and Chen, Junshen K and Chuang, Jo and V{\'a}zquez, Marynel and Savarese, Silvio},
  booktitle={Proceedings of the IEEE/CVF Conference on Computer Vision and Pattern Recognition},
  pages={11276--11286},
  year={2021}
}

@inproceedings{raychaudhuri2021law,
  title={Language-aligned waypoint (law) supervision for vision-and-language navigation in continuous environments},
  author={Raychaudhuri, Sonia and Wani, Saim and Patel, Shivansh and Jain, Unnat and Chang, Angel},
  booktitle={Proceedings of the 2021 conference on empirical methods in natural language processing},
  pages={4018--4028},
  year={2021}
}

@inproceedings{georgakis2022cross,
  title={Cross-modal map learning for vision and language navigation},
  author={Georgakis, Georgios and Schmeckpeper, Karl and Wanchoo, Karan and Dan, Soham and Miltsakaki, Eleni and Roth, Dan and Daniilidis, Kostas},
  booktitle={Proceedings of the IEEE/CVF conference on computer vision and pattern recognition},
  pages={15460--15470},
  year={2022}
}

@article{chen2022weakly,
  title={Weakly-supervised multi-granularity map learning for vision-and-language navigation},
  author={Chen, Peihao and Ji, Dongyu and Lin, Kunyang and Zeng, Runhao and Li, Thomas and Tan, Mingkui and Gan, Chuang},
  journal={Advances in Neural Information Processing Systems},
  volume={35},
  pages={38149--38161},
  year={2022}
}

@article{wang2024sim,
  title={Sim-to-real transfer via 3d feature fields for vision-and-language navigation},
  author={Wang, Zihan and Li, Xiangyang and Yang, Jiahao and Liu, Yeqi and Jiang, Shuqiang},
  journal={arXiv preprint arXiv:2406.09798},
  year={2024}
}

@inproceedings{zhang2025mapnav,
  title={Mapnav: A novel memory representation via annotated semantic maps for vlm-based vision-and-language navigation},
  author={Zhang, Lingfeng and Hao, Xiaoshuai and Xu, Qinwen and Zhang, Qiang and Zhang, Xinyao and Wang, Pengwei and Zhang, Jing and Wang, Zhongyuan and Zhang, Shanghang and Xu, Renjing},
  booktitle={Proceedings of the 63rd Annual Meeting of the Association for Computational Linguistics (Volume 1: Long Papers)},
  pages={13032--13056},
  year={2025}
}

@inproceedings{tong2024eyes,
  title={Eyes wide shut? exploring the visual shortcomings of multimodal llms},
  author={Tong, Shengbang and Liu, Zhuang and Zhai, Yuexiang and Ma, Yi and LeCun, Yann and Xie, Saining},
  booktitle={Proceedings of the IEEE/CVF conference on computer vision and pattern recognition},
  pages={9568--9578},
  year={2024}
}

@inproceedings{du2024embspatial,
  title={Embspatial-bench: Benchmarking spatial understanding for embodied tasks with large vision-language models},
  author={Du, Mengfei and Wu, Binhao and Li, Zejun and Huang, Xuan-Jing and Wei, Zhongyu},
  booktitle={Proceedings of the 62nd Annual Meeting of the Association for Computational Linguistics (Volume 2: Short Papers)},
  pages={346--355},
  year={2024}
}

@article{stogiannidis2025mind,
  title={Mind the gap: Benchmarking spatial reasoning in vision-language models},
  author={Stogiannidis, Ilias and McDonagh, Steven and Tsaftaris, Sotirios A},
  journal={arXiv preprint arXiv:2503.19707},
  year={2025}
}

@inproceedings{chen2024mapgpt,
  title={Mapgpt: Map-guided prompting with adaptive path planning for vision-and-language navigation},
  author={Chen, Jiaqi and Lin, Bingqian and Xu, Ran and Chai, Zhenhua and Liang, Xiaodan and Wong, Kwan-Yee},
  booktitle={Proceedings of the 62nd Annual Meeting of the Association for Computational Linguistics (Volume 1: Long Papers)},
  pages={9796--9810},
  year={2024}
}

@article{li2026trajectory,
  title={Trajectory-Diversity-Driven Robust Vision-and-Language Navigation},
  author={Li, Jiangyang and Wan, Cong and Dong, SongLin and Ding, Chenhao and Wang, Qiang and Ma, Zhiheng and Gong, Yihong},
  journal={arXiv preprint arXiv:2603.15370},
  year={2026}
}

@article{mcinnes2018umap,
  title={Umap: Uniform manifold approximation and projection for dimension reduction},
  author={McInnes, Leland and Healy, John and Melville, James},
  journal={arXiv preprint arXiv:1802.03426},
  year={2018}
}

@article{Qwen2.5-VL,
  title={Qwen2.5-VL Technical Report},
  author={Bai, Shuai and Chen, Keqin and Liu, Xuejing and Wang, Jialin and Ge, Wenbin and Song, Sibo and Dang, Kai and Wang, Peng and Wang, Shijie and Tang, Jun and Zhong, Humen and Zhu, Yuanzhi and Yang, Mingkun and Li, Zhaohai and Wan, Jianqiang and Wang, Pengfei and Ding, Wei and Fu, Zheren and Xu, Yiheng and Ye, Jiabo and Zhang, Xi and Xie, Tianbao and Cheng, Zesen and Zhang, Hang and Yang, Zhibo and Xu, Haiyang and Lin, Junyang},
  journal={arXiv preprint arXiv:2502.13923},
  year={2025}
}

@article{zhang2025flatland,
  title={From flatland to space: Teaching vision-language models to perceive and reason in 3d},
  author={Zhang, Jiahui and Chen, Yurui and Zhou, Yanpeng and Xu, Yueming and Huang, Ze and Mei, Jilin and Chen, Junhui and Yuan, Yu-Jie and Cai, Xinyue and Huang, Guowei and others},
  journal={arXiv preprint arXiv:2503.22976},
  year={2025}
}

@article{liu2026span,
  title={SPAN-Nav: Generalized Spatial Awareness for Versatile Vision-Language Navigation},
  author={Liu, Jiahang and Xu, Tianyu and Chen, Jiawei and Yue, Lu and Zhang, Jiazhao and Wang, Zhiyong and Li, Minghan and Zhao, Qisheng and Li, Anqi and Su, Qi and others},
  journal={arXiv preprint arXiv:2603.09163},
  year={2026}
}

@article{wang2026vlingnav,
  title={VLingNav: Embodied Navigation with Adaptive Reasoning and Visual-Assisted Linguistic Memory},
  author={Wang, Shaoan and Luo, Yuanfei and Chen, Xingyu and Luo, Aocheng and Li, Dongyue and Liu, Chang and Chen, Sheng and Zhang, Yangang and Yu, Junzhi},
  journal={arXiv preprint arXiv:2601.08665},
  year={2026}
}

@article{li2026one,
  title={One Agent to Guide Them All: Empowering MLLMs for Vision-and-Language Navigation via Explicit World Representation},
  author={Li, Zerui and Zheng, Hongpei and Zhao, Fangguo and Chan, Aidan and Zhou, Jian and Lin, Sihao and Li, Shijie and Wu, Qi},
  journal={arXiv preprint arXiv:2602.15400},
  year={2026}
}

@article{habibpour2025think,
  title={Think, Remember, Navigate: Zero-Shot Object-Goal Navigation with VLM-Powered Reasoning},
  author={Habibpour, Mobin and Afghah, Fatemeh},
  journal={arXiv preprint arXiv:2511.08942},
  year={2025}
}

@inproceedings{hudson2019gqa,
  title={Gqa: A new dataset for real-world visual reasoning and compositional question answering},
  author={Hudson, Drew A and Manning, Christopher D},
  booktitle={Proceedings of the IEEE/CVF conference on computer vision and pattern recognition},
  pages={6700--6709},
  year={2019}
}

@article{zhao2024imaginenav,
  title={Imaginenav: Prompting vision-language models as embodied navigator through scene imagination},
  author={Zhao, Xinxin and Cai, Wenzhe and Tang, Likun and Wang, Teng},
  journal={arXiv preprint arXiv:2410.09874},
  year={2024}
}

@article{wang2025dreamnav,
  title={Dreamnav: A trajectory-based imaginative framework for zero-shot vision-and-language navigation},
  author={Wang, Yunheng and Fang, Yuetong and Wang, Taowen and Feng, Yixiao and Tan, Yawen and Zhang, Shuning and Liu, Peiran and Ji, Yiding and Xu, Renjing},
  journal={arXiv preprint arXiv:2509.11197},
  year={2025}
}

@article{wang2026imaginenav++,
  title={ImagineNav++: Prompting Vision-Language Models as Embodied Navigator through Scene Imagination},
  author={Wang, Teng and Zhao, Xinxin and Cai, Wenzhe and Sun, Changyin},
  journal={IEEE Transactions on Pattern Analysis and Machine Intelligence},
  year={2026},
  publisher={IEEE}
}

\clearpage
\beginappendix
\section{Detailed Reward Formulations}
\label{app:rewards}

\subsection{Local Step Rewards}
Based on our algorithm implementation, we define five unscaled reward components $r_i^{(t)}$ for a given state $s_t$ and action $a_t$:

\begin{enumerate}
    \item \textbf{Target Approach Reward ($r_{app}^{(t)}$)}: Let $d_t$ be the expert geodesic distance.
We calculate the distance reduction $\Delta d = d_t - d_{t+1}$ and the reward is then mapped via a signed sigmoid:
    \begin{equation}
        r_{app}^{(t)} = \frac{2}{1 + \exp(-\Delta d)} - 1
    \end{equation}
    \item \textbf{Collision Avoidance Reward ($r_{coll}^{(t)}$)}: Based on the distance to the nearest obstacle $c_t$, which can be directly queried from the Habitat simulator, we define a clamped reward:
    \begin{equation}
        r_{coll}^{(t)} = \max(0, \min(c_t, 1.0))
    \end{equation}
    
    \item \textbf{Stop Action Reward ($r_{stop}^{(t)}$)}: It encourages stopping near the goal while penalizing premature stops or moving away from the goal in the vicinity of the 
target:
    \begin{equation}
        r_{stop}^{(t)} = \begin{cases} 
        1.0, & \text{if } a_t = \text{STOP and } d_t < 1.5 \\ 
        -1.0, & \text{if } a_t = \text{STOP and } d_t > 3.0 \\ 
        0.5, & \text{if } 1.5 <= d_t < 2.5 \text{ and } (a_t = \text{STOP or } \Delta d >= 0) \\ 
        -0.5, & \text{if } 1.5 <= d_t < 2.5 
\text{ and } \Delta d < 0 \\ 
        0.0, & \text{otherwise} 
        \end{cases}
    \end{equation}
    
\item \textbf{Reasoning Value Reward ($r_{rea}^{(t)}$)}: This component evaluates the specific contribution of Reasoning (CoT) to the current navigation progress. We define the reward based on the \textbf{performance margin} of the selected action's approach reward $r_{app}^{(t)}$ over the local baseline. Let $\bar{r}_{app}^{(t)} = \frac{1}{M}\sum_{j=1}^M r_{app}^{(t, j)}$ be the mean approach reward across all $M$ sampled candidates at step $t$. The reward is formulated as:
\begin{equation}
    r_{rea}^{(t)} = h_t \cdot \big(r_{app}^{(t)} - \bar{r}_{app}^{(t)}\big)
\end{equation}
where $h_t \in \{0, 1\}$ is the CoT indicator at step $t$. This design ensures that the model is rewarded only when its reasoning leads to a decision that outperforms the average local policy, effectively penalizing redundant or misleading CoT.
    
    \item \textbf{Format Adherence Reward ($r_{fmt}^{(t)}$)}: Ensures the model's output strictly follows the structured JSON 
format and valid spatial constraints:
    \begin{equation}
        r_{fmt}^{(t)} = \mathbb{I}_{valid} \cdot \rho_{bounds}
    \end{equation}
    where $\mathbb{I}_{valid} \in \{0, 1\}$ indicates whether the parsed JSON matches the required schema, and $\rho_{bounds} = 0.0$ if the predicted pixel $(u, v)$ falls outside the valid sensor range $[0, 279]^2$, otherwise $1.0$.
\end{enumerate}

\subsection{Global Trajectory Rewards}
The global reward evaluates the holistic quality of the completed trajectory, consisting of three components:

\begin{enumerate}
    \item \textbf{Task Success Reward ($r_{suc}$)}: The primary driver for navigation completion, relaxing strict success to oracle boundaries:
    \begin{equation}
        r_{suc} = \begin{cases} 
        1.0, & \text{if strictly successful} \\ 
        0.5, & \text{if oracle successful} \\ 
        0.0, & \text{otherwise} 
        \end{cases}
    \end{equation}
    
    \item \textbf{Trajectory Efficiency Reward ($r_{spl}$)}: Optimizes navigation efficiency using Success weighted by Path Length (SPL), rewarding optimal trajectory execution:
    \begin{equation}
        r_{spl} = \text{SPL}
    \end{equation}
    
    \item \textbf{Reasoning Density Reward ($r_{den}$)}: Enforces a reasoning density constraint to prevent over-thinking.
Let $r$ be the ratio of reasoning steps in the sequence.
    \begin{equation}
        r_{den} = \begin{cases}  
        1.0 - r, & \text{if } 0 \le r \le 0.4 \\ 
        0.6 - 3.0(r - 0.4), & \text{if } 0.4 < r \le 0.6 \\ 
        0.0, & \text{if } r > 0.6 
        \end{cases}
    \end{equation}

\end{enumerate}

\subsection{Loss Formulas}
\label{subsec:loss_formulas}

In our framework, the output at each decision step $t$ (comprising Chain-of-Thought reasoning and action prediction) is a sequence of tokens. Let the generated sequence for the $g$-th trajectory at step $t$ be $a_{g,t} = \{y_1, y_2, \dots, y_K\}$, where $K$ is the number of tokens. our policy update is computed strictly at the \textbf{token level}. 

Let $\rho_{g,t,k}$ denote the policy ratio for the $k$-th token at time step $t$:
\begin{equation}
    \rho_{g,t,k} = \frac{\pi_{\theta}(y_k \,|\, C_{g,t}, y_{<k})}{\pi_{\theta_{old}}(y_k \,|\, C_{g,t}, y_{<k})}
\end{equation}
where $C_{g,t}$ represents the historical context (including the initial instruction and previous observations), and $y_{<k}$ denotes the tokens generated before $y_k$ within the current step.

The policy is optimized via the token-level GRPO objective. The step-level advantage $A_S^{(g,t)}$ is broadcast to all constituent tokens within that step:
\begin{equation}
\begin{split}
    L_{GRPO}(\theta) = \mathbb{E}\Bigg[\frac{1}{G}\sum_{g=1}^{G} \sum_{t=1}^{T_g} \frac{1}{K} \sum_{k=1}^{K} \bigg( & \min\Big(\rho_{g,t,k} A_{S}^{(g,t)}, \text{clip}(\rho_{g,t,k}, 1-\epsilon, 1+\epsilon)A_{S}^{(g,t)}\Big) \\
    & - \beta D_{KL}\big(\pi_{\theta}(\cdot|C_{g,t}, y_{<k}) \,||\, \pi_{ref}(\cdot|C_{g,t}, y_{<k})\big)\bigg)\Bigg]
\end{split}
\end{equation}
where $T_g$ is the total number of decision steps in the $g$-th trajectory, and $A_S^{(g,t)}$ is the superimposed advantage (combining global and local rewards) corresponding to that specific step.

\subsection{Hyperparameter Settings}
\label{app:hyperparameters}

In our implementation of the Two-Level GRPO framework, the hyperparameters for the reward formulations and the annealed guided sampling strategy are empirically set as follows:

\textbf{Local Step Reward Weights}: To balance fine-grained execution and reasoning, the weights are set to $\lambda_1 = 0.5$ (Target Approach), $\lambda_2 = 0.2$ (Collision Avoidance), $\lambda_3 = 0.8$ (Stop Action), $\lambda_4 = 0.5$ (Reasoning Value), and $\lambda_5 = 0.2$ (Format Adherence).
    
\textbf{Global Trajectory Reward Weights}: To strongly prioritize ultimate navigation task completion while maintaining trajectory efficiency and reasoning sparsity, the weights are set to $\omega_1 = 3.0$ (Task Success), $\omega_2 = 1.0$ (Trajectory Efficiency), and $\omega_3 = 1.0$ (Reasoning Density).

\textbf{Annealed Guided Sampling}: For the guidance coefficient $\beta_k = \beta_0 \cdot \alpha^k$, we set the initial value $\beta_0 = 2.0$ to ensure high-quality early rollouts and an exponential decay rate $\alpha = 0.99$ to gradually transition towards uniform exploration.

\section{Related Work}
\label{sec:related_work}

\subsection{Action Formulation in VLM-based Navigation}

Recent VLM-based navigators typically force the model to output predefined textual atomic actions (\cite{zhang2024navid},\cite{cheng2024navila},\cite{wei2025streamvln}) or regress continuous 3D spatial coordinates (\cite{zhang2025embodied}). However, recent analytical studies and benchmarks in multimodal perception expose a systematic representation gap: because foundational VLMs rely heavily on 2D contrastive vision encoders (e.g., CLIP) and image-text pairs, they suffer from profound shortcomings in explicit 3D spatial and geometric reasoning (\cite{stogiannidis2025mind, tong2024eyes, du2024embspatial}). Forcing these 2D-native models to predict unnatural 3D coordinates or rigid atomic steps directly contradicts their pre-training priors, often resulting in severe geometric hallucinations and poor sample efficiency. To mitigate this, recent literature argues that tightly coupled designs severely limit system performance, advocating for the decoupling of high-level semantic planning from low-level spatial execution (\cite{li2026one}). However, this method requires the continuous, asynchronous construction of 3D metric and topological maps, incurring massive additional overhead. Other decoupling approaches attempt to simplify the VLM's role into a multiple-choice selection by pre-computing navigable trajectories or employing diffusion models to generate future observations(\cite{zhao2024imaginenav,wang2025dreamnav,wang2026imaginenav++}). However, this not only introduces substantial computational overhead but also restricts the agent's exploration to predefined heuristic proposals.

Sharing this decoupled philosophy, \textbf{TAMP-Nav} introduces a \textbf{Pixel-to-3D} action space. By prompting the VLM to act merely as an intuitive 2D ``pointer'' in the image plane, projecting the selected pixel into a 3D spatial coordinate, and offloading low-level execution to a deterministic SLAM controller, our paradigm seamlessly aligns embodied execution with the VLM's inherent 2D visual capabilities, largely bypassing the 2D-to-3D representation gap.
\subsection{Reasoning and Memory in VLM-based Navigation}

Existing navigation methods often struggle to balance reasoning depth with memory efficiency over long horizons. On one hand, methods incorporating Chain-of-Thought (CoT) (\cite{song2023llmplanner, lin2025navcot, liu2026span, habibpour2025think}) enforce rigid, dense reasoning at every step, causing linear surges in inference latency. While a recent method, VLingNav (\cite{wang2026vlingnav}), proposes adaptive reasoning to address this issue, it relies on coarse data annotations and lacks an in-depth exploration of how to systematically inject this meta-cognitive capability into the model. Simultaneously, memory designs like compressing historical frames into sparse tokens (\cite{zhang2024navid}) or relying on text-based maps (\cite{chen2024mapgpt}) inevitably lead to topological discontinuity or modal information loss. Although recent state-of-the-art models such as StreamVLN (\cite{wei2025streamvln}) and NavFoM (\cite{zhang2025embodied}) have introduced 3D spatial voxelization or implicit forgetting curves to handle memory budgets, these compression and forgetting mechanisms may inadvertently discard critical historical details. Furthermore, they fundamentally lack explicit logical reasoning capabilities, resulting in significant performance degradation on complex, long-horizon tasks.

\textbf{TAMP-Nav} resolves these dilemmas via \textbf{Selective Reasoning and Anchor-Trajectory Memory}. By triggering deep CoT reasoning only at critical decision nodes and compressing routine traversed paths into lightweight spatiotemporal tokens, our framework strictly bounds the computational budget while maintaining global topological awareness and logical coherence across long horizons.
\subsection{Reinforcement Learning for VLM-based Navigation}
To alleviate the covariate shift in Imitation Learning, recent works apply GRPO to embodied navigation.
However, they face two extremes: either relying exclusively on extreme sparse trajectory-level rewards (\cite{li2026trajectory}), which exacerbates credit assignment difficulties;
or over-relying on strict expert action matching (\cite{qi2025vln}), which restricts open-ended exploration.
\textbf{TAMP-Nav} breaks this bottleneck with a \textbf{Two-Level GRPO} paradigm, superimposing global terminal advantages with local step advantages to provide dense supervision without relying on step-by-step expert-forcing.

\section{Algorithm for Key Node Selection}
\label{sec:appendix_key_node}

To ensure the reasoning process is only triggered at critical waypoints rather than at every step, we utilize a Spatiotemporal Key Node Mining strategy during the construction of the MultiNav-CoT dataset. The detailed algorithmic process is summarized in Algorithm \ref{alg:key_node_selection}. It consists of visual-semantic importance scoring, distance-based greedy filtering to maintain reasoning sparsity, and temporal in-filling to preserve global topological connectivity. In our implementation, we empirically set the spatial thresholds $D_{min} = 2$ and $D_{max} = 6$ to balance node sparsity with trajectory coverage.

To ensure a balanced contribution from both modalities in the combined importance score $S(t) = S_{sem}(t) + S_{vis}(t)$, we explicitly map both the semantic relevance and visual transition metrics to the strict $[0, 1]$ interval via trajectory-level Min-Max normalization. Let $E_{text}(\cdot)$ and $E_{vis}(\cdot)$ denote the pre-trained CLIP text and image encoders, respectively.

\textbf{Semantic Importance Score ($S_{sem}$)}: 
The raw semantic score measures the cosine similarity between the instruction $Q$ and the current visual observation $v_t$. The normalized score $S_{sem}(t)$ is defined as:
\begin{equation}
    s_{sem}^{raw}(t) = \frac{E_{text}(Q) \cdot E_{vis}(v_t)}{\|E_{text}(Q)\| \|E_{vis}(v_t)\|}
\end{equation}
\begin{equation}
    S_{sem}(t) = \frac{s_{sem}^{raw}(t) - \min\limits_{j \in [1,T]} s_{sem}^{raw}(j)}{\max\limits_{j \in [1,T]} s_{sem}^{raw}(j) - \min\limits_{j \in [1,T]} s_{sem}^{raw}(j)}
\end{equation}

\textbf{Visual Transition Score ($S_{vis}$)}:
To capture significant scene transitions (e.g., turning a corner or entering a new room), the raw visual score measures the feature difference (calculated as $1 - \text{cosine similarity}$) between the current frame $v_t$ and the previous frame $v_{t-1}$. For the first frame ($t=1$), $s_{vis}^{raw}(1)$ is defined as $0$. The normalized score $S_{vis}(t)$ is formulated as:
\begin{equation}
    s_{vis}^{raw}(t) = 1 - \frac{E_{vis}(v_t) \cdot E_{vis}(v_{t-1})}{\|E_{vis}(v_t)\| \|E_{vis}(v_{t-1})\|}
\end{equation}
\begin{equation}
    S_{vis}(t) = \frac{s_{vis}^{raw}(t) - \min\limits_{j \in [1,T]} s_{vis}^{raw}(j)}{\max\limits_{j \in [1,T]} s_{vis}^{raw}(j) - \min\limits_{j \in [1,T]} s_{vis}^{raw}(j)}
\end{equation}

By independently standardizing $S_{sem}(t)$ and $S_{vis}(t)$ within each trajectory, they inherently share identical scales, ensuring a balanced contribution to the final selection objective $S(t)$.

\begin{algorithm}[h]
   \caption{Key Node Selection Strategy}
   \label{alg:key_node_selection}
\begin{algorithmic}
   \STATE {\bfseries Input:} Trajectory $\mathcal{T}$ (with spatial poses $p_t$ and visual frames $v_t$), Instruction $Q$, Spatial Thresholds $D_{min}, D_{max}$
   \STATE {\bfseries Output:} Key Node Set $\mathcal{K}$

   \STATE \textit{// 1. Importance Scoring (Vectorized \& Normalized)}
   \STATE Compute $S(t) \leftarrow S_{sem}(t) + S_{vis}(t)$ for all $t \in [1, T]$
   
   \STATE \textit{// 2. Distance-based Greedy Filtering}
   \STATE $\mathcal{I} \leftarrow \text{Argsort}(S)[0 : 0.3T]$ \COMMENT{Top-30\% indices descending}
   \STATE $\mathcal{K} \leftarrow \emptyset$
   \FOR{$t \in \mathcal{I}$}
      \IF{$\mathcal{K} = \emptyset \lor \min_{k \in \mathcal{K}} \text{dist}_{spatial}(p_t, p_k) \ge D_{min}$}
         \STATE $\mathcal{K} \leftarrow \mathcal{K} \cup \{t\}$
      \ENDIF
   \ENDFOR

   \STATE \textit{// 3. Temporal In-filling (Padding)}
   \STATE Sort $\mathcal{K}$ by time index
   \WHILE{$\exists \text{ adjacent pair } (t_{curr}, t_{next}) \text{ in } \mathcal{K} \text{ s.t. } \text{dist}_{spatial}(p_{t_{curr}}, p_{t_{next}}) > D_{max}$}
      \STATE $\mathcal{K} \leftarrow \mathcal{K} \cup \{\arg\max_{t \in (t_{curr}, t_{next})} S(t)\}$
      \STATE Sort $\mathcal{K}$ by time index
   \ENDWHILE

   \STATE \textbf{return} $\mathcal{K}$
\end{algorithmic}
\end{algorithm}

\section{Real-World Experiment Details}
\label{sec:appendix_real_world}

\begin{figure}[htbp]
    \centering
    
\hfill
    \begin{subfigure}[b]{0.34\textwidth}
        \centering
        \includegraphics[width=\textwidth]{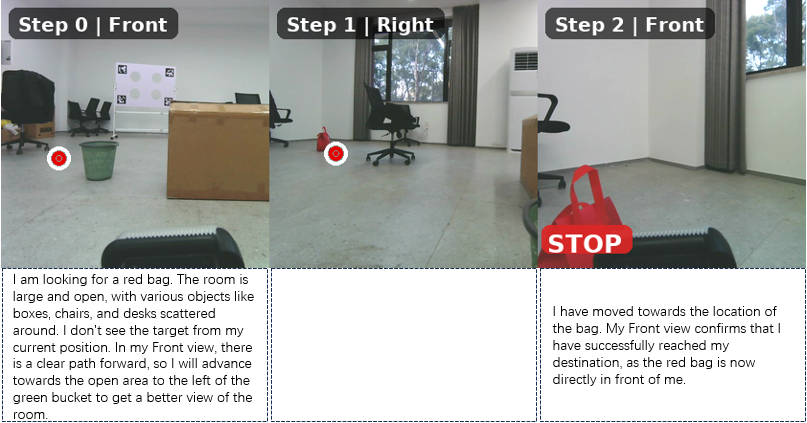}
        \caption{Instruction: \textit{``Find the red bag''}}
        \label{fig:real_task_bag}
    \end{subfigure}
    \hfill \begin{subfigure}[b]{0.48\textwidth}
        \centering
        \includegraphics[width=\textwidth]{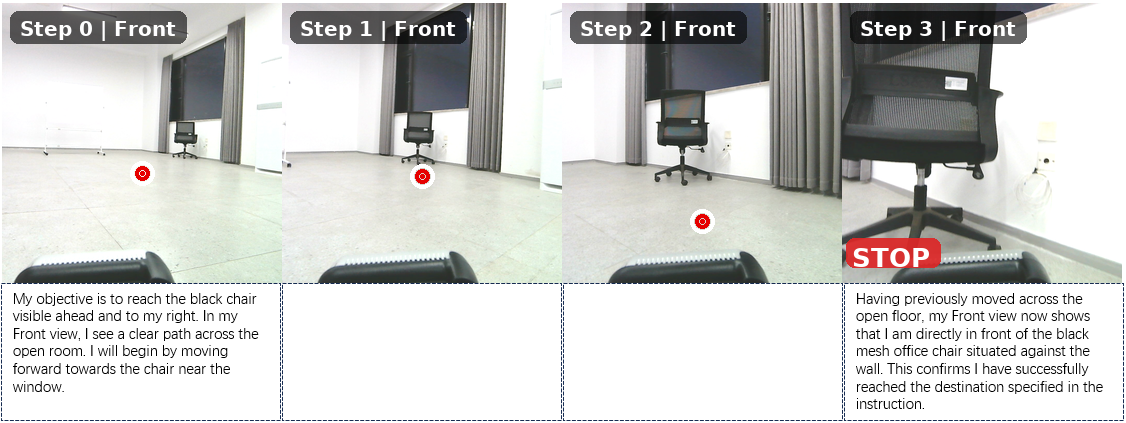}
        \caption{Instruction: \textit{``Maps to the black chair on the right.''}}
        \label{fig:real_task_chair}
    \end{subfigure}
    \hfill
    
\begin{subfigure}[b]{0.82\textwidth}
        \centering
        \includegraphics[width=\textwidth]{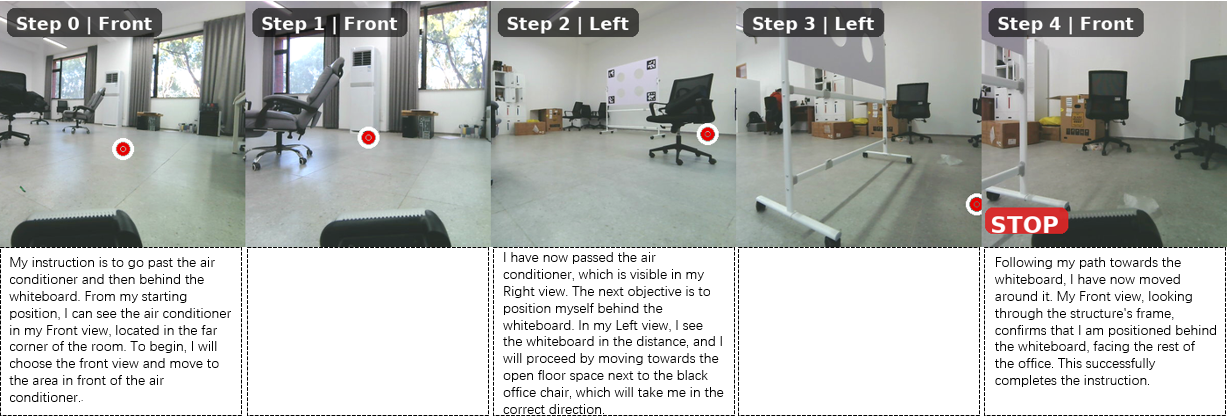}
        \caption{Instruction: \textit{``Proceed past the air conditioner and navigate behind the whiteboard.''}}
        \label{fig:real_task_whiteboard}
    \end{subfigure}
    
\begin{subfigure}[b]{0.82\textwidth}
        \centering
        \includegraphics[width=\textwidth]{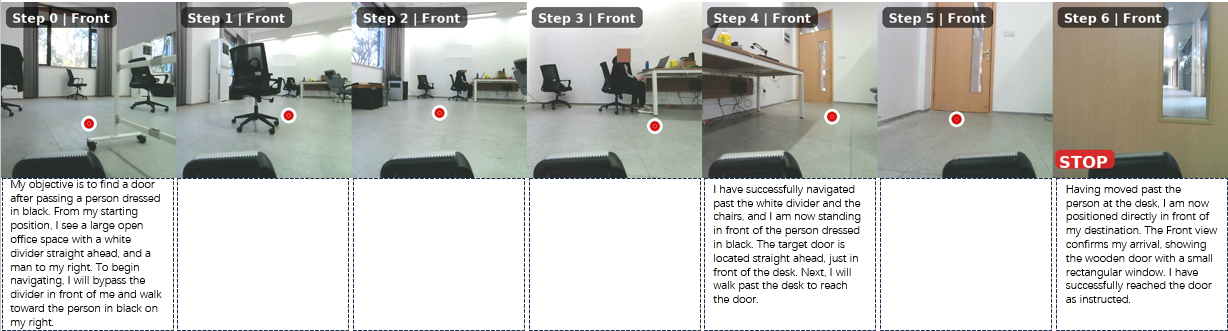}
        \caption{Instruction: \textit{``Walk past the person in black and reach the door.''}}
        \label{fig:real_task_door}
    \end{subfigure}
    
\begin{subfigure}[b]{0.82\textwidth}
        \centering
        \includegraphics[width=\textwidth]{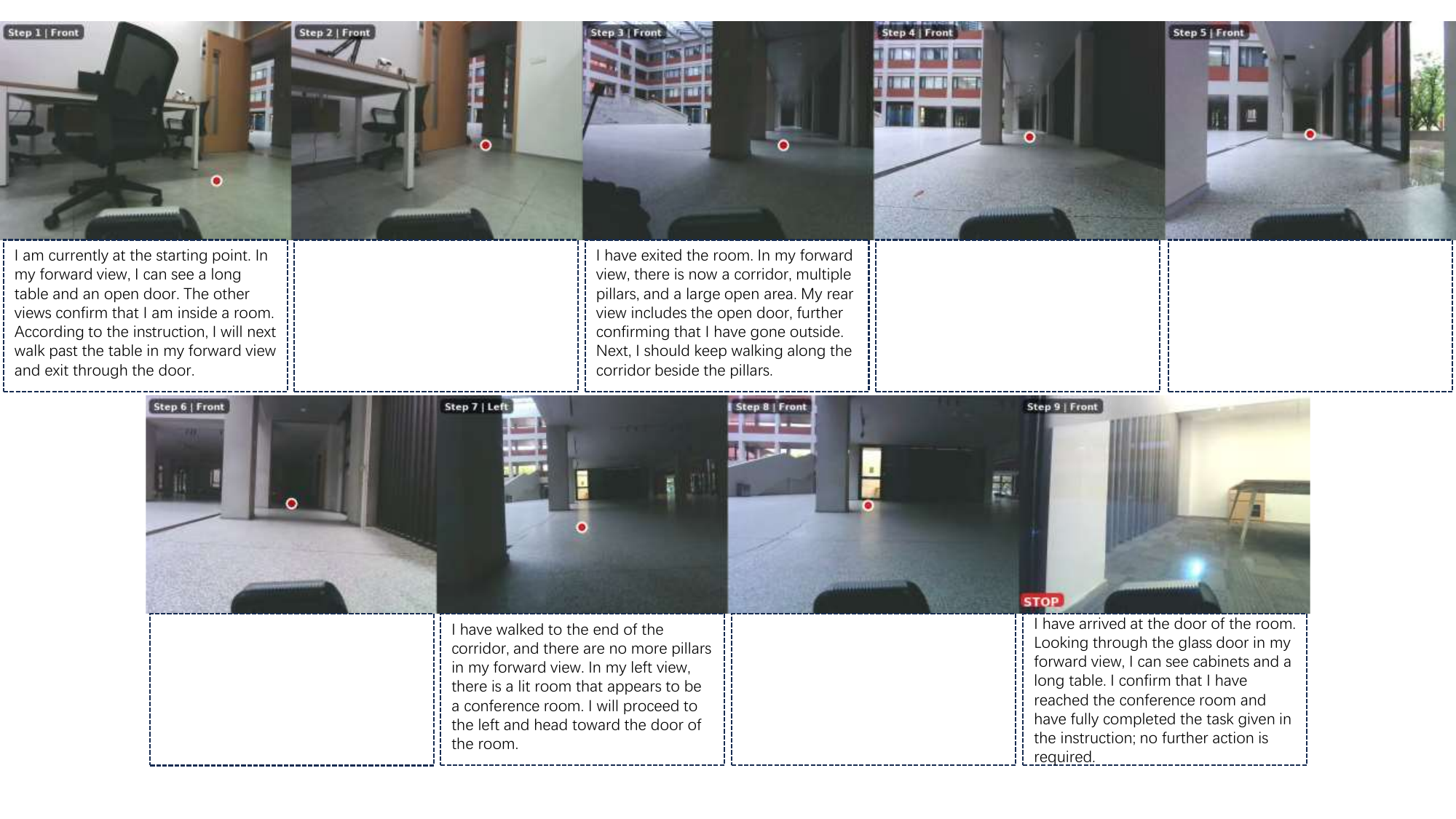}
        \caption{Instruction: \textit{``Go out the door, walk along the corridor next to the pillars to the end, then turn left and go to the entrance of the lit conference room.''}}
        \label{fig:real_task_outdoor}
    \end{subfigure}
    
    \caption{\textbf{Examples of real-world experiments.} The agent demonstrates zero-shot transferability across physical environments, executing multi-turn instructions by reasoning and outputting sequential 2D pixel coordinates.}
    \label{fig:real_world_examples}
\end{figure}

\textbf{Hardware and System Setup.} Our deployment platform is a Unitree Go2 quadruped robot, equipped with 4 annularly distributed RGB-D cameras (each providing an approximately $100^\circ$ Field of View to collectively ensure $360^\circ$ coverage) and a Hesai XT16 LiDAR mounted underneath. To bridge the sim-to-real gap, the system architecture adopts a decoupled design:

(1) \textbf{High-Level Perception}: TAMP-Nav runs on a remote server, processing real-time multi-view images from the cameras and generating 2D pixel coordinates as navigation targets.

(2) \textbf{Low-Level Execution}: The system utilizes the aligned depth measurements from the selected RGB-D camera to directly project the VLM-selected pixels into 3D world coordinates. It employs FastLIO (\cite{xu2022fastlio2}) for ego-motion estimation and utilizes FAR Planner (\cite{yang2022far}) for local path planning and dynamic obstacle avoidance.

This decoupled Pixel-to-3D point selection design ensures that the VLM focuses on high-level semantic localization and reasoning, while intrinsic odometry, path execution, and motion safety are delegated to a mature classical SLAM stack.

\textbf{Qualitative Analysis in Unseen Environments.} Consistent with the evaluation protocol detailed in Section 4.6, we tested our framework in unmapped physical spaces featuring diverse objects, corridors, and obstacles. Figure \ref{fig:real_world_examples} illustrates five representative execution trajectories. The agent demonstrates accurate object localization (Figure \ref{fig:real_world_examples}\subref{fig:real_task_bag} and \ref{fig:real_world_examples}\subref{fig:real_task_chair}) and effectively follows sequential instructions involving spatial relationships (Figure \ref{fig:real_world_examples}\subref{fig:real_task_whiteboard} and \ref{fig:real_world_examples}\subref{fig:real_task_door}). Furthermore, Figure \ref{fig:real_world_examples}\subref{fig:real_task_outdoor} showcases an extended navigation task, confirming the model's capability to process complex, multi-stage instructions over longer physical distances. As noted in the main text, TAMP-Nav is trained exclusively in simulation without real-world fine-tuning. The behavior in these unseen physical environments supports the zero-shot sim-to-real feasibility of our proposed paradigm.

\section{Analysis of STI token}
\label{sec:appendix_STI_token}
\begin{figure}[h]
    \centering
    \includegraphics[width=0.75\textwidth]{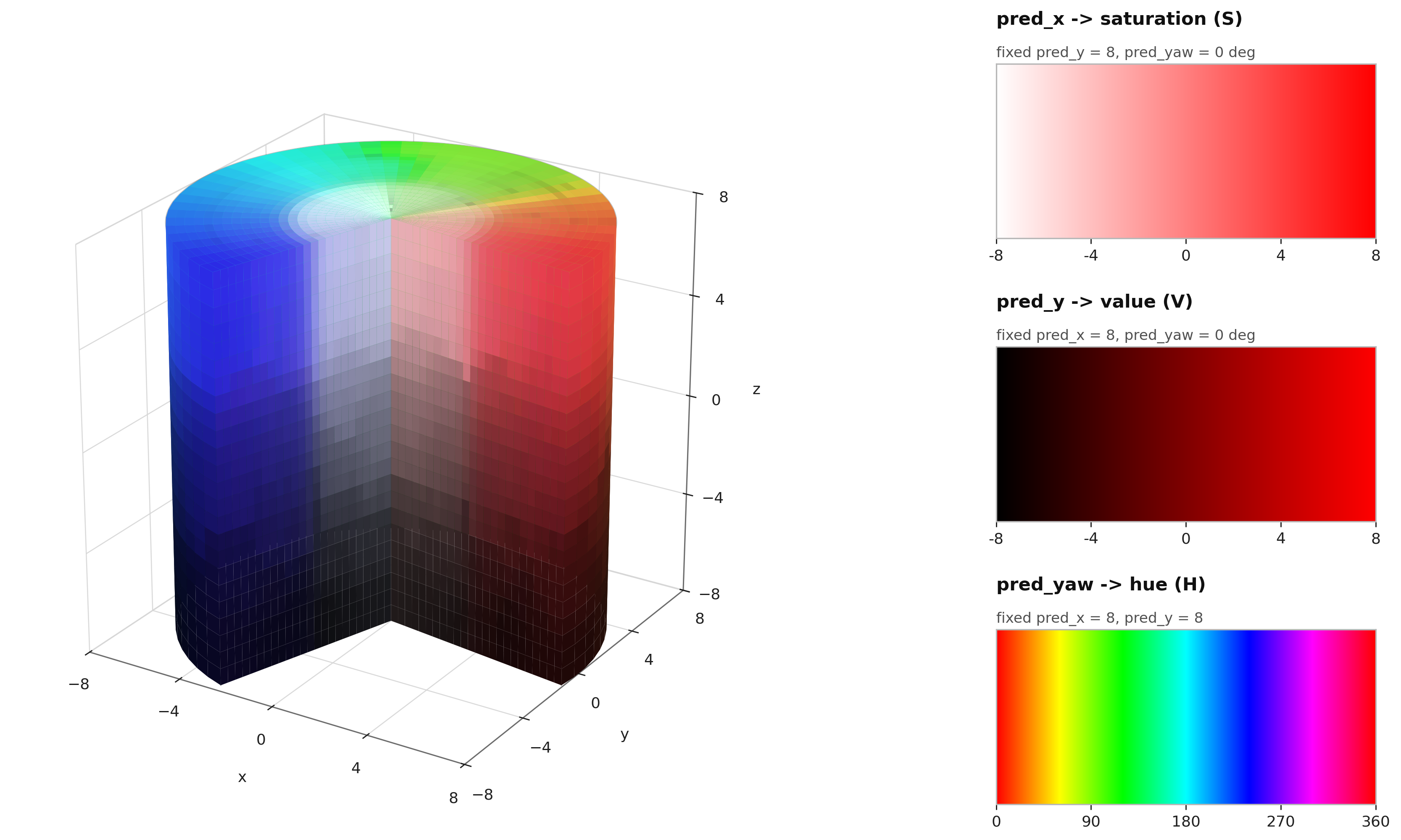}
    \caption{Visualization of STI token. We employ a clustering algorithm (\cite{mcinnes2018umap}) to map high-dimensional embeddings into a 3D space.}
    \label{fig:STI}
\end{figure}
To verify whether the trained positional encoder effectively retains pose information, we input the initial pose $p=(x,y,\mathrm{yaw})$ into the positional encoder $f_{\text{pos}}$ to obtain the corresponding STI token embedding:
\[e=f_{\text{pos}}(x,y,\mathrm{yaw})\in\mathbb{R}^d\]
where $d$ denotes the output dimension of the encoder. In our implementation, this embedding serves as the unified pose representation processed by the subsequent VLM. 

To visualize whether these high-dimensional embeddings preserve the underlying geometric structure of the original physical poses, we perform supervised dimensionality reduction. Specifically, we first randomly sample a large-scale set of poses and compute their corresponding embeddings. We then employ a clustering algorithm (\cite{mcinnes2018umap}) to compress these embeddings into a 4-dimensional latent variable:
\[z=(z_x,z_y,z_s,z_c)\]
which is optimized to align with the normalized ground-truth components:
\[\left(\frac{x}{R},\frac{y}{R},\sin(\mathrm{yaw}),\cos(\mathrm{yaw})\right)\]
where $R$ is a constant representing the spatial range. Subsequently, the predicted pose $(\hat x,\hat y,\hat{\mathrm{yaw}})$ is reconstructed from the latent variable $(z_x,z_y,z_s,z_c)$.

As illustrated in Figure \ref{fig:STI}, we design a visualization scheme analogous to the HSV color space to simultaneously represent the three pose variables. In this geometric coordinate system, the circumferential direction corresponds to $\mathrm{yaw}$, the radial distance from the center corresponds to $x$, and the vertical height corresponds to $y$. Consequently, the spatial position of each block in the figure is determined by its original ground-truth pose. 

In parallel, the color of each block is mapped from its reconstructed pose: Hue ($H$) corresponds to $\hat{\mathrm{yaw}}$, Saturation ($S$) to $\hat x$, and Value ($V$) to $\hat y$. If the positional encoder successfully preserves the pose information, the reconstructed color field should transition consistently with the spatial layout: hue should shift smoothly along the circumference, saturation along the radius, and value along the vertical axis. This visualization demonstrates that the "ground-truth poses define spatial positions, while poses recovered from embeddings define colors"; the high degree of alignment between the two confirms the fidelity of our positional encoder.
\section{Simulation Execution Example}
\label{sec:appendix_sim_example}

\begin{figure}[htbp]
    \centering
    \includegraphics[width=0.95\textwidth]{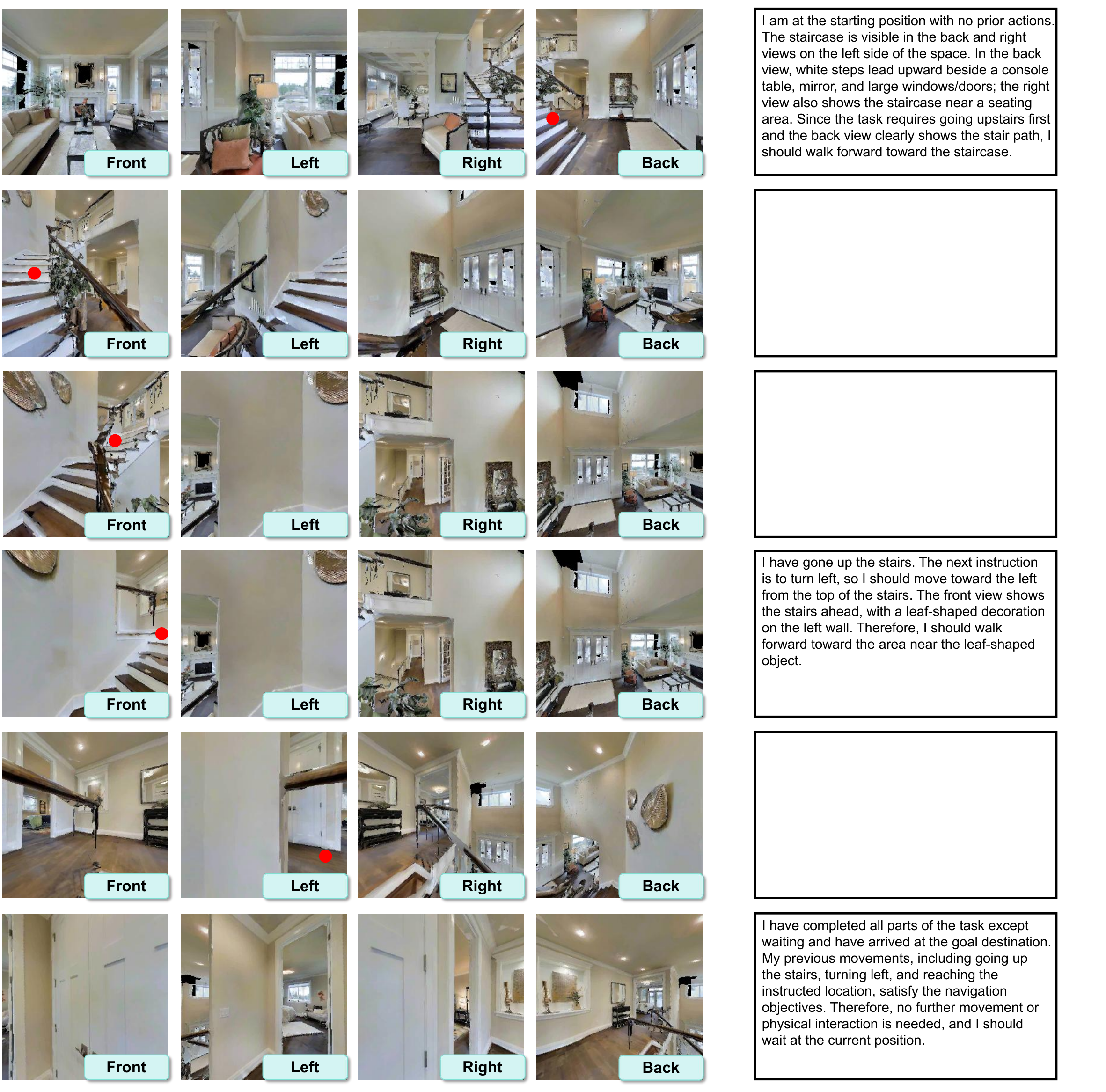}
    \caption{\textbf{Visualization of a simulated navigation trajectory.} The agent successfully executes the complex instruction: \textit{``Go up the stairs and turn left. Wait at the doorway to the bedroom straight ahead.''} by triggering selective reasoning at key decision points and outputting point-and-click actions.}
    \label{fig:sim_example}
\end{figure}

To provide a more intuitive understanding of TAMP-Nav's decision-making process, Figure \ref{fig:sim_example} illustrates a representative navigation trajectory in the simulation environment. 

Given the multi-step instruction: \textit{``Go up the stairs and turn left. Wait at the doorway to the bedroom straight ahead.''}, the agent demonstrates robust visual-semantic grounding and logical planning. As shown in the figure, the model autonomously triggers its reasoning mechanism (CoT) at critical topological nodes-such as initially locating the stairs, determining the correct turning direction upon reaching the top floor, and confidently executing the stop action once the final destination is reached. This qualitative example highlights the effectiveness of our reasoning-on-demand and pixel-level point selection paradigm.

\section{Prompts for Structured CoT Generation}
\label{sec:appendix_cot_prompts}

To enhance reasoning quality and mitigate hallucinations, TAMP-Nav employs a multi-stage prompt engineering pipeline. The tables below detail the core prompts utilized at each stage during the construction of the MultiNav-CoT dataset. \textbf{Please note that the content presented is a textually refined version. It is designed to clearly illustrate the core cognitive guidance and reasoning rules imposed on the model, and thus omits the complex JSON formatting validations and defensive directives found in the original engineering scripts.}

\begin{table}[htbp]
\caption{Pre-processing: Instruction Quality Filtering}
\centering
\small
\begin{tabular}{p{0.95\textwidth}}
\toprule
\textbf{System Prompt} \\
You are a strict filter for navigation instructions. \\
- Pure action = a movement-only command without a target object/landmark/destination, e.g., "go straight", "turn left". \\
- Not pure action = includes a target or landmark, e.g., "go straight to the bed". \\
If there are 4 or more consecutive pure actions anywhere in the instruction, output DROP. Otherwise output KEEP. Consecutive means back-to-back pure actions with no non-pure-action instruction between them. \\
\midrule
\textbf{User Prompt Template} \\
Instruction: \{task\} \\
\bottomrule
\end{tabular}
\end{table}

\begin{table}[htbp]
\caption{Stage 1: Task Phase Localization (History Summary)}
\centering
\small
\begin{tabular}{p{0.95\textwidth}}
\toprule
\textbf{System Prompt} \\
You are an expert at summarizing navigation trajectories. Given a task and image sequences, your job is to summarize the trajectory from image $0 \rightarrow k$ in ONE continuous paragraph. Explicitly state which part of the task has been completed. Do NOT mention future parts. Return ONLY a JSON object. \\
\midrule
\textbf{User Prompt Template} \\
Task: \{task\} \\
Summary ranges: \{segment\_list\_text\} \\
For EACH range [0 $\rightarrow$ k], write ONE paragraph in first person. Describe what happened and which part of the task is completed by that point. \\
\bottomrule
\end{tabular}
\end{table}

\begin{table}[htbp]
\caption{Stage 2: Current Observation Analysis}
\centering
\small
\begin{tabular}{p{0.95\textwidth}}
\toprule
\textbf{System Prompt} \\
You are a controlled reasoning module. Describe the indoor scene layout (open paths, obstacles, landmarks) based on the attached front, left, back, and right views. Only describe visible objects. Do NOT mention targets, goals, history, or next steps. \\
\midrule
\textbf{Output Format} \\
\texttt{"Analyze the current observations": [your detailed description]} \\
\bottomrule
\end{tabular}
\end{table}

\begin{table}[htbp]
\caption{Stage 3: Future Action Reasoning (Infer Next Step)}
\centering
\small
\begin{tabular}{p{0.95\textwidth}}
\toprule
\textbf{System Prompt} \\
\textbf{IRON RULE:} IF \texttt{gt\_action == "stop"}, declare arrival. ELSE, reason step-by-step how the move advances the task. The final sentence MUST name the target view and an object-defined region. \textbf{FORBIDDEN:} coordinates, pixels, red circles, proportions. \\
\midrule
\textbf{User Prompt Template} \\
Task: \{task\} $|$ History: \{summary\} $|$ Analysis: \{analysis\} \\
Target Guidance: \{target\_guidance\} (Use this to locate the region, but do NOT mention the guidance itself). \\
\bottomrule
\end{tabular}
\end{table}

\begin{table}[htbp]
\caption{Stage 4: CoT Fusion and Polish}
\centering
\small
\begin{tabular}{p{0.95\textwidth}}
\toprule
\textbf{System Prompt} \\
Rewrite the three-stage reasoning into a single, coherent English paragraph. Preserve spatial logic and the target view. Remove redundancies. Do NOT include stage titles. Output ONLY the polished paragraph. \\
\bottomrule
\end{tabular}
\end{table}

\section{Detailed Training Configurations}
\label{sec:appendix_training_configs}

In this section, we provide the comprehensive training hyperparameters and experimental configurations for the Supervised Fine-Tuning (SFT) and the Two-Level GRPO stages of \textbf{TAMP-Nav}. All experiments were conducted on a computing cluster equipped with \textbf{8 NVIDIA A800 GPUs}, each with \textbf{80GB of VRAM}. Regarding computational cost, the Supervised Fine-Tuning (SFT) phase required approximately \textbf{160 GPU hours} (around 20 wall-clock hours) to complete, while the Two-Level GRPO phase took approximately \textbf{600 GPU hours} (around 75 wall-clock hours).

\subsection{Supervised Fine-Tuning (SFT) Phase}
During the SFT cold-start phase, we fine-tune the base Qwen2.5-VL-7B model to internalize the core navigation logic. To mitigate catastrophic forgetting of the model's inherent general visual reasoning capabilities, we incorporate approximately 10\% of \texttt{gqa\_interleaved\_cot} into our MultiNav-CoT training set. This dataset is a custom variant derived from the original GQA dataset (\cite{hudson2019gqa}), augmented with interleaved Chain-of-Thought (CoT) annotations to align with our proposed paradigm.

The SFT is conducted for 1 epoch using the Bfloat16 (bf16) precision format. To optimize memory efficiency, we employ Gradient Checkpointing and Flash Attention 2. We utilize the AdamW optimizer with a cosine learning rate scheduler and a warmup ratio of 0.03. The peak learning rate is set to $5 \times 10^{-6}$, with a weight decay of 0.01 and a maximum gradient norm of 5.0. The maximum sequence length is configured to 4096 tokens. The per-device training batch size is 1, with gradients accumulated over 32 steps to achieve a larger effective batch size.

\subsection{Two-Level GRPO Phase}
Following SFT, we apply the Two-Level GRPO framework to align the agent's reasoning with physical execution. For the environmental setups and navigation instructions, we directly reuse the training dataset from the SFT phase. During this phase, the group size for trajectory-level rollouts is set to $G = 8$, and the candidate set size for step-level rollouts is $M = 4$. The maximum rollout trajectory length is capped at 24 steps.

For text generation during environmental exploration, we set the sampling temperature to $0.7$ and top-p to $0.9$ to encourage diverse trajectory sampling. The maximum prompt length is 4096 tokens, and the maximum completion length is 512 tokens. In the GRPO loss formulation, the KL divergence penalty coefficient ($\beta$) is set to $0.0$. 

The reinforcement learning (RL) phase is conducted for 800 training steps using bf16 precision. We continue to use the AdamW optimizer with a cosine scheduler, but with a reduced peak learning rate of $2 \times 10^{-6}$. The per-device training batch size is set to 2, with a gradient accumulation step of 32.

\end{document}